\documentclass[sigconf, nonacm]{acmart}
\usepackage{balance}
\usepackage{makecell}
\usepackage{multirow}
\usepackage{enumitem}
\usepackage{graphicx}
\usepackage{subfigure}
\usepackage[normalem]{ulem}
\usepackage{url}
\usepackage{enumerate}
\usepackage{booktabs}
\usepackage{multirow}
\usepackage{xspace}

\usepackage{amsmath}
\usepackage{color}
\usepackage{threeparttable}
\usepackage{ulem}
\usepackage{amssymb}

\usepackage{float}
\usepackage[ruled,linesnumbered]{algorithm2e}
\usepackage{algorithm2e, setspace, algpseudocode}

\def\bb#1{\mathbb{#1}}
\def\b#1{\mathbf{#1}}
\def\cal#1{\mathcal{#1}}

\newcommand{\eg}{\textit{e.g.,}\xspace}
\newcommand{\ie}{\textit{i.e.,}\xspace}

\newtheorem{definition}{Definition}

\newcommand\vldbpagestyle{empty}
\newcommand\vldbdoi{10.14778/3551793.3551827}
\newcommand\vldbpages{2733 - 2746}
\newcommand\vldbvolume{15}
\newcommand\vldbissue{11}
\newcommand\vldbyear{2022}
\newcommand\vldbauthors{\authors}
\newcommand\vldbtitle{\shorttitle} 
\newcommand\vldbavailabilityurl{URL_TO_YOUR_ARTIFACTS}

\title{Decoupled Dynamic Spatial-Temporal Graph Neural Network for Traffic Forecasting}

\begin{document}

\author{Zezhi Shao$^{1,2}$, Zhao Zhang$^{1}$, Wei Wei$^{3, *}$, Fei Wang$^{1, *}$, Yongjun Xu$^{1}$, Xin Cao$^{4}$, Christian S. Jensen$^{5}$}
\affiliation{%
 \institution{$^1$Institute of Computing Technology, Chinese Academy of Sciences, Beijing, China\\
 $^2$University of Chinese Academy of Sciences, Beijing, China\\
 $^3$School of Computer Science and Technology, Huazhong University of Science and Technology, Wuhan, China\\
 $^4$School of Computer Science and Engineering, The University of New South Wales, Australia\\
 $^5$Department of Computer Science, Aalborg University, Denmark\\
 \{shaozezhi19b, zhaozhang2021, wangfei, xyj\}@ict.ac.cn, weiw@hust.edu.cn, xin.cao@unsw.edu.au, csj@cs.aau.dk}\country{}}

\renewcommand{\vldbauthors}{Zezhi Shao, Zhao Zhang, Wei Wei, Fei Wang, Yongjun Xu, Xin Cao, and Christian S. Jensen}
\begin{abstract}
We all depend on mobility, and vehicular transportation affects the daily lives of most of us.
Thus, the ability to forecast the state of traffic in a road network is an important functionality and {\color{black}a challenging task}.
Traffic data is {\color{black}often} obtained from sensors deployed in a road network. 
{\color{black}Recent proposals on spatial-temporal graph neural networks have achieved great progress at modeling complex spatial-temporal correlations in traffic data, by modeling traffic data as a diffusion process. 
However, intuitively, traffic data encompasses two different kinds of hidden time series signals, namely the diffusion signals and inherent signals.
Unfortunately, nearly all previous works coarsely consider traffic signals entirely as the outcome of the diffusion, while neglecting the inherent signals, which impacts model performance negatively.}
To improve modeling performance, {\color{black}we propose a novel \textbf{D}ecoupled \textbf{S}patial-\textbf{T}emporal \textbf{F}ramework (DSTF) that separates the diffusion and inherent traffic information in a data-driven manner, which encompasses a unique estimation gate and a residual decomposition mechanism.  
The separated signals can be handled subsequently by the diffusion and inherent modules separately.
Further, we propose an instantiation of DSTF, \textbf{D}ecoupled \textbf{D}ynamic \textbf{S}patial-\textbf{T}emporal \textbf{G}raph \textbf{N}eural \textbf{N}etwork~(D$^2$STGNN), that} captures spatial-temporal {\color{black}correlations} and also features a dynamic graph learning module that targets the learning of the dynamic characteristics of traffic networks. 
Extensive experiments with four real-world traffic datasets demonstrate that the framework is capable of advancing the state-of-the-art.
\end{abstract}

\maketitle

\pagestyle{\vldbpagestyle}
\begingroup\small\noindent\raggedright\textbf{PVLDB Reference Format:}\\
\vldbauthors. \vldbtitle. PVLDB, \vldbvolume(\vldbissue): \vldbpages, \vldbyear.\\
\href{https://doi.org/\vldbdoi}{doi:\vldbdoi}
\endgroup
\begingroup
\renewcommand\thefootnote{}\footnote{\noindent
* Corresponding author. \\
This work is licensed under the Creative Commons BY-NC-ND 4.0 International License. Visit \url{https://creativecommons.org/licenses/by-nc-nd/4.0/} to view a copy of this license. For any use beyond those covered by this license, obtain permission by emailing \href{mailto:info@vldb.org}{info@vldb.org}. Copyright is held by the owner/author(s). Publication rights licensed to the VLDB Endowment. \\
\raggedright Proceedings of the VLDB Endowment, Vol. \vldbvolume, No. \vldbissue\ %
ISSN 2150-8097. \\
\href{https://doi.org/\vldbdoi}{doi:\vldbdoi} \\
}\addtocounter{footnote}{-1}\endgroup

\ifdefempty{\vldbavailabilityurl}{}{
\vspace{.3cm}
\begingroup\small\noindent\raggedright\textbf{PVLDB Artifact Availability:}\\
The source code of this research paper has been made publicly available at \url{https://github.com/zezhishao/D2STGNN}.
\endgroup
}
\section{Introduction}
\label{Section1}

\begin{figure}[t]
  \centering
  \includegraphics[width=1\linewidth]{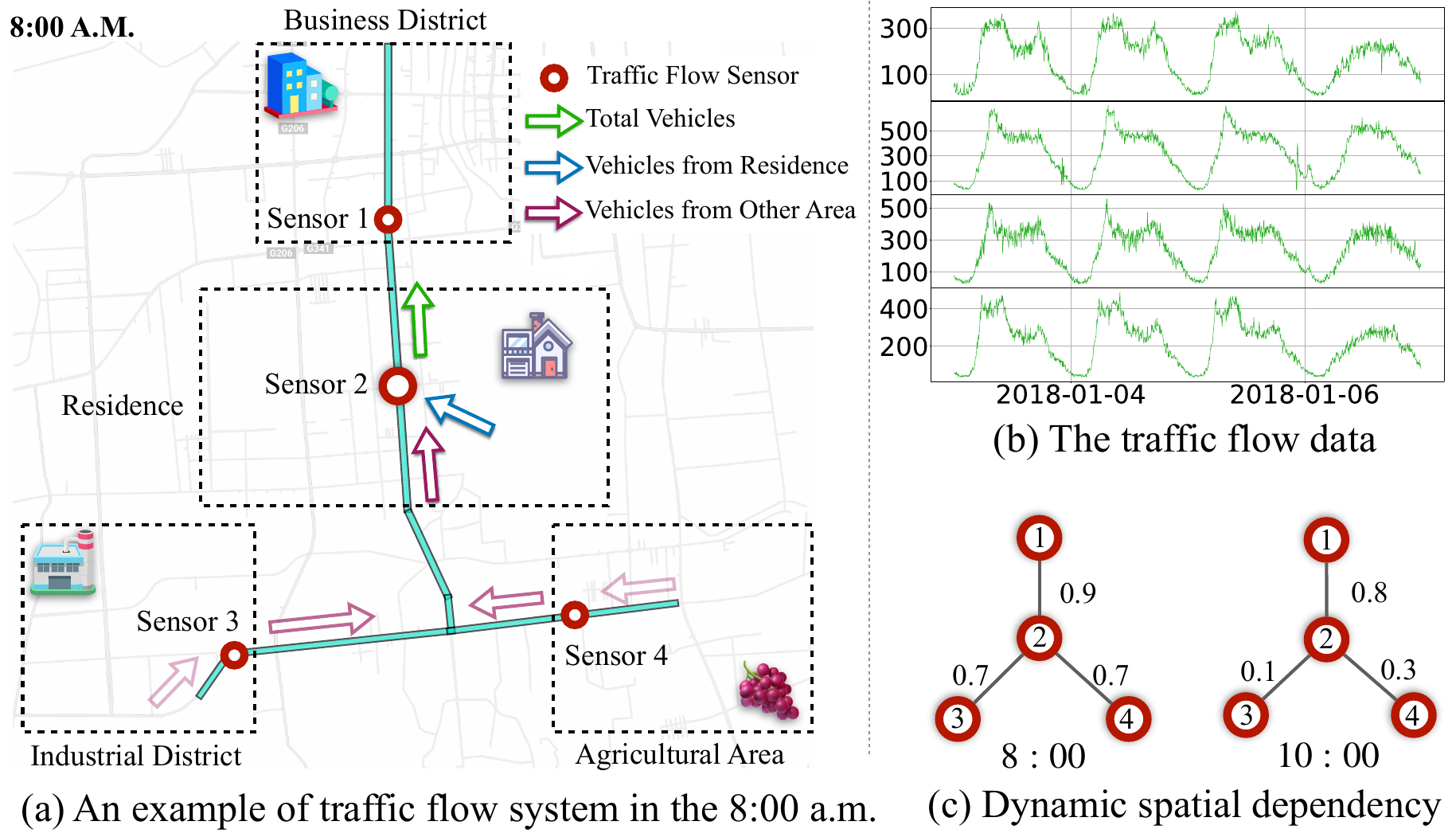}
  \caption{\color{black}An example of the traffic flow system.
  }
  \label{Intro1}
\end{figure}

{\color{black}
Traffic forecasting is a crucial service in Intelligent Transportation Systems~(ITS) to predict future traffic conditions (\eg traffic flow\footnote{{\color{black}An example of a traffic flow system is shown in Figure \ref{Intro1}(a), where traffic sensors are deployed at important locations in the road network and record the total number of vehicles passing during a unit time interval. 
Over time, we can get four time series corresponding to sensors 1 to 4, as shown in Figure \ref{Intro1}(b).}})
based on historical traffic conditions~\cite{2020GMAN} observed by sensors~\cite{ICDE21S,VLDB21MDTP}.
This functionality fuels a wide range of services related to traffic management~\cite{TrafficManagement}, urban computing~\cite{UrbanComputing}, public safety~\cite{PublicSafty}, and beyond~\cite{incident, Demand, Travel, qian2020cabin}.

Previous traffic forecasting studies usually fall into two categories, \ie knowledge-driven~\cite{cascetta2013transportation} and data-driven~\cite{ARIMAKarman1,2014GRU,2017DCRNN,GWNet}.
The former commonly adopt queuing theory for user behavior simulation in traffic~\cite{cascetta2013transportation}, while neglecting the natural complexity of real-world traffic flow.
Regarding the latter, many early studies formulate the problem as a simple time series (\eg single variant time series) prediction task~\cite{2021DGCRN}
and address it via various conventional statistic-based methods, such as auto-regressive integrated moving average (ARIMA~\cite{ARIMA}) and Kalman filtering~\cite{ARIMAKarman1}.
These methods do not handle the high non-linearity of each time series well,
since they typically rely heavily on stationarity-related assumptions.
More importantly, they disregard the complex correlations among time series, which severely limits the effectiveness of traffic forecasting.

Recently, deep learning-based approaches~\cite{lv2014traffic, 2018DMVST, 2017DCRNN} have been proposed to capture the complex spatial-temporal correlations in traffic flow. 
A promising and effective way is to construct an adjacency matrix to model the complex spatial topology of a road network and formulates the traffic data as a spatial-temporal graph.
An example is shown in Figure \ref{Intro}(a), where each node represents a sensor, and the signals on each node vary over time.
Sequentially, STGNN-based methods are proposed for traffic forecasting that models the dynamics of the traffic flow as a diffusion process~\cite{2017DCRNN, GWNet}, and combines diffusion graph convolution~\cite{2017DCRNN} and sequential models~\cite{2016TCN, 2014GRU} to jointly model complex spatial-temporal correlations.
The former~\cite{2017DCRNN} models the diffusion of vehicles among sensors in a road network, \ie the spatial dependency.
The latter~\cite{2016TCN, 2014GRU} models the temporal dynamics, \ie the temporal dependency.
\color{black}
\begin{figure}[t]
  \centering
  \includegraphics[width=0.929\linewidth]{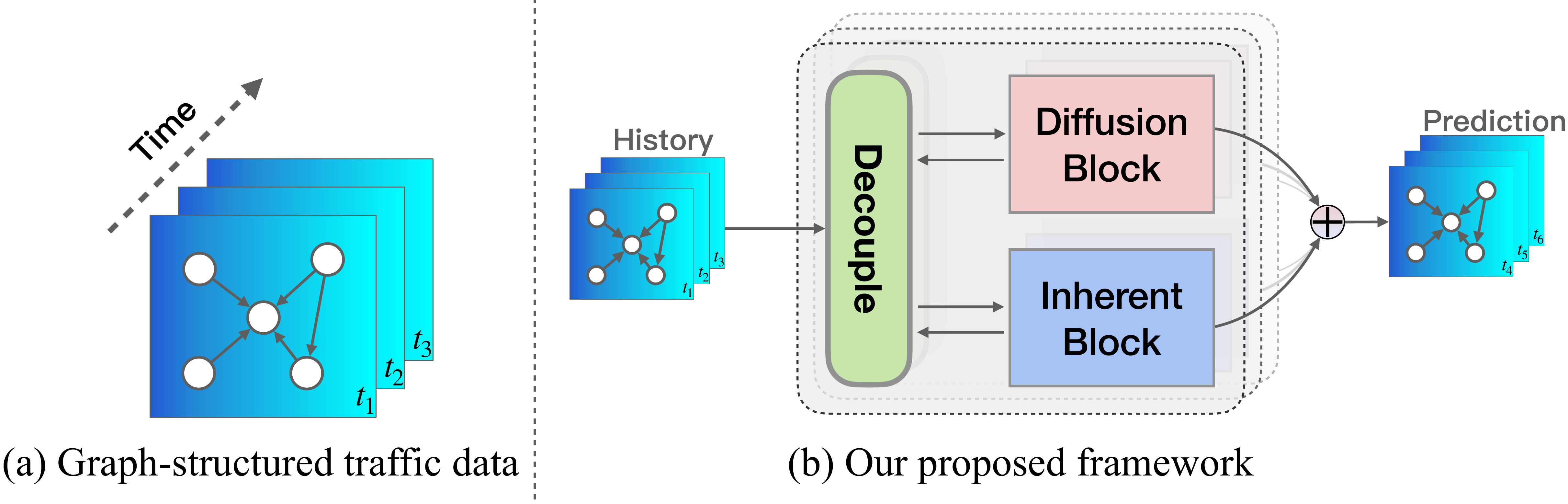}
  \caption{
  {\color{black}Graph structured traffic data and our proposed framework.}
  }
  \label{Intro}
\end{figure}

Although encouraging results have been achieved, these methods still fail to fully exploit the complex spatial-temporal correlations. 
First of all, each signal~(\ie time series) naturally contains two different types of signals, \ie diffusion and non-diffusion signals (which is also called inherent signal for simplicity).
The diffusion signal captures the vehicles diffused from other sensors, while the non-diffusion signal captures the vehicles that are independent of other sensors.
However, almost all previous studies consider traffic data as a diffusion signal while disregarding the non-diffusion signal.
That is, they model the complex spatial-temporal correlations coarsely.
However, a reasonable solution is to exploit the complex spatial-temporal correlations more subtly,
\ie explicitly modeling the diffusion and inherent signal simultaneously.
Second, the pre-defined adjacency matrix in STGNN-based methods is static, and thus such construction methods may severely restrict the representative ability to complex road networks, making it difficult for these methods to model the dynamics of traffic flow.
We illustrate them with examples in Figure \ref{Intro1}. 
Without loss of generality, Figure \ref{Intro1} presents a typical traffic flow system.
Important locations in the road network are equipped with traffic sensors that record traffic flow data, \ie the number of vehicles during a unit time interval.
From Figure \ref{Intro1}, we make two observations.
\textbf{(I)} The recorded values of each sensor are affected by two factors, \ie the diffusion signal and the non-diffusion signal.
As shown in Figure \ref{Intro1}(a), vehicles passing through sensor 2~(green arrow) at 8 a.m. come from two parts.
The first part is vehicles that depart directly from somewhere in the area near the sensor~(blue arrow), \eg vehicles that drive directly from residence to the business district to work.
The other part is vehicles diffused from adjacent areas~(wine-red arrow), \eg vehicles that drive from the industrial district (sensor 3) and the agricultural area (sensor 4) to provide daily supplies.
The former is independent of other sensors, while the latter is an artifact of the diffusion process.
We call them \textit{hidden inherent time series} and \textit{hidden diffusion time series}, respectively, and each time series in Figure \ref{Intro1}(b) is a superposition of them.
\textbf{(II)} The traffic flow within the same road network may change over time, \ie spatial dependency is dynamic. 
An example is shown in Figure \ref{Intro1}(c), where the traffic at sensors 3 and 4 can significantly affect sensor 2 at 8 a.m., while there is only a small influence at 10 a.m. 

Therefore, addressing the above issues to effectively leverage all complex spatial-temporal correlations in traffic data is essential for improving the performance of traffic forecasting. 
To achieve this, we first propose a \textbf{D}ecoupled \textbf{S}patial-\textbf{T}emporal \textbf{F}ramework (DSTF) that is illustrated in the diagram in Figure \ref{Intro}(b).
DSTF separates the diffusion and inherent traffic information using a the decouple block in a data-driven manner.
Furthermore, we design a dynamic graph learning module based on a self-attention mechanism to address the second issue.
The above designs are key elements of an instantiation of DSTF, called the \textbf{D}ecoupled \textbf{D}ynamic \textbf{S}patial-\textbf{T}emporal \textbf{G}raph \textbf{N}eural \textbf{N}etwork~(D$^2$STGNN).
Specifically, we first design the decouple block shown in Figure \ref{D2GNN}, which contains a residual decomposition mechanism and an estimation gate to decompose traffic data. 
The former removes the parts of signals that the diffusion and inherent models can approximate well. Thus, the parts of signals that are not learned well is retained. 
The latter estimates roughly the proportion of the two kinds of signals to relieve the burden of the first model in each layer, which takes the original signal as input and needs to learn specific parts in it.
Second, the dynamic graph learning module comprehensively exploits available information to adjust the road network-based spatial dependency by learning latent correlations between time series based on the self-attention mechanism.
In addition, specialized diffusion and inherent models, for the two hidden time series are designed according to their particular characteristics.
A spatial-temporal localized convolution is designed to model the \textit{hidden diffusion time series}.
A recurrent neural network and self-attention mechanism are used jointly to model the \textit{hidden inherent time series}.}

In summary, the main contributions are the following:

\begin{itemize}
    \item 
    {\color{black}We propose a novel Decoupled Spatial-Temporal Framework (DSTF) for traffic forecasting, which decouples the hidden time series generated by the diffusion process and the hidden time series that is independent of other sensors.
    This enables more precise modeling of the different parts of traffic data to improve prediction accuracy.
    \item
    Based on the DSTF, a dynamic graph learning module is proposed that takes into account the dynamic nature of spatial dependency.
    Besides, we design a diffusion model and a inherent model to handle the two hidden time series.
    The above design forms our instantiation of DSTF, D$^2$STGNN.}
    
    \item We conduct extensive experiments on four real-world, large-scale datasets to gain insight into the effectiveness of {\color{black}the framework DSTF and the instantiation D$^2$STGNN}. 
    Experimental results show that our proposal is able to consistently {\color{black}and significantly} outperforms all baselines.
\end{itemize}

The paper is organized as follows.
Section \ref{Section2} covers related work, and Section \ref{Section3} presents preliminaries and the problem definition.
In Section \ref{Section4}, we present the decoupled spatial-temporal framework in detail.
Section \ref{Section5} details the chosen instantiation of the framework, D$^2$STGNN.
We {\color{black}present extensive performance experiments and prediction visualizations in Section \ref{Section6}. 
We also report on extensive ablation studies of different architectures, important components, and training strategies.}
Section \ref{Section7} concludes the paper.
\section{Related Work}
\label{Section2}

{\color{black}With the availability of large-scale traffic data and the rise of artificial intelligence~\cite{Innovation}, Spatial-Temporal Graph Neural Networks~(STGNNs) are proposed to model the complex spatial-temporal correlations in traffic data~\cite{2017DCRNN, 2018GaAN, 2018STGCN, 2019STMetaNet, shao2022pre}.
Generally speaking, STGNNs model the traffic system as a diffusion process~\cite{2017DCRNN} and combine the diffusion convolutions~\cite{2017DCRNN, GWNet} and sequential models~\cite{2014GRU, 2016TCN} to jointly model the spatial-temporal correlation.
The diffusion convolutions are variants of Graph Convolution Networks~(GCN~\cite{2013FirstGCN, 2016ChebNet, 2017GCN}), which are well suited to deal with the non-euclidean relationships between multiple time series in traffic data.
Sequential models, such as GRU~\cite{2014GRU}, LSTM~\cite{FC-LSTM}, and TCN~\cite{2016TCN}, are used to model temporal dependency.
For example, DCRNN~\cite{2017DCRNN} integrates diffusion convolution and the sequence to sequence architecture~\cite{2014Seq2Seq} to model the diffusion process.
Graph WaveNet~\cite{GWNet} combines diffusion convolution with dilated casual convolution~\cite{2016TCN} to capture spatial-temporal correlation efﬁciently and effectively.

Recent works focus on designing more powerful diffusion convolution models and sequential models. 
For example, many variants of GCNs, such as GAT~\cite{2018GAT}, MixHop~\cite{2019MixHop}, and SGC~\cite{2019SGC}, are adapted to STGNNs for better performance~\cite{2020MTGNN, 2020CGC, 2019STGRAT, 2020STSGCN, 2020TSSRGCN, 2020SLCNN}.
The attention mechanism~\cite{2017Transformer} and its variants, which theoretically have infinite receptive field size, are widely used to capture long-term temporal dependencies~\cite{2021ASTGNN} in the sequential model~\cite{2020STGNN, Informer}.
Moreover, a few very recent works propose to model the dynamic spatial dependency~\cite{2020GMAN, 2021DMSTGCN}.
The idea is to learn the latent correlations between nodes based on dynamic node feature, which is usually represented by the combination of real-time traffic features and other external features.
For example, 
GMAN~\cite{2020GMAN} designs a spatial attention mechanism by considering traffic features and node embeddings from the graph structure to learn the attention score.

Although STGNNs have made considerable progress, we find there is still significant room for improvement.
Firstly, existing works solely consider the traffic data as a diffusion signal while neglecting the non-diffusion one, as discussed in Section \ref{Intro}.
They model the complex spatial-temporal correlations in a coarse manner, which may impact model performance negatively.
Secondly, although there are a few works on modeling dynamic spatial dependency, they do not consider all available information.
Most of them explore the dynamic spatial dependency based on the feature of traffic conditions, ignoring either the constraints of the road network topology~\cite{2021DMSTGCN}, or the time~\cite{2020GMAN} or node~\cite{2019ASTGCN} information.
}
\section{Preliminaries}
\begin{table}
    \caption{Frequently used notation. }
    \label{tab:notations}
    \begin{tabular}{m{1.3cm}<{\centering}|m{6.3cm}}
      \toprule
      \textbf{Notations} & \textbf{Definitions}\\
       \midrule
      $G$ & The traffic network $G=(V, E)$ with node set $V$ and edge set $E$. \\
      $N$ & Number of sensors~(nodes) of the traffic network, \ie $|V|=N$. \\
      $\b{A}$ & The adjacency matrix of traffic network $G$. \\
      $T_h$ & The number of past traffic signals considered. \\
      $T_p$ & The number of future time steps to forecast. \\
      $C$ &  Number of feature channels in a traffic signal. \\
      $d$ & Dimensionality of hidden states.\\
      $\mathbf{E}$ & Embedding of the sensors~(nodes).\\
      $\mathbf{T}$ & Embedding of the time steps.\\
      $\mathbf{W}$ & Parameter matrix of the fully connected layer.\\
      $\mathbf{X}_t$ & Traffic signal at the $t$-th time step. \\
      $\mathbf{H}_t$ & Hidden state at the $t$-th time step. \\
      $\mathcal{X}$ & Traffic signals of the $T_h$ most recent past time steps. \\
      $\mathcal{Y}$ & Traffic signals of the $T_f$ nearest-future time steps. \\
      $\mathcal{H}$ & Hidden states over multiple time steps.\\
      $\odot$ & Element-wise product.\\
      $\parallel$ & Concatenation.\\
      $\mathit{Concat}(\cdot)$ & Broadcast concatenation.\\
      \bottomrule
    \end{tabular}
  \end{table}
\label{Section3}
We first define the notions of traffic network and traffic signal, and then define the forecasting problem addressed. 
Frequently used notations are summarized in Table \ref{tab:notations}.
\begin{definition}
\textbf{Traffic Sensor.} 
A traffic sensor is a sensor deployed in a traffic system, such as a road network, and it records traffic information such as the flow of passing vehicles or vehicle speeds.
\end{definition}
\begin{definition}
\textbf{Traffic Network.} 
A traffic network is a directed or undirected graph $G=(V, E)$, where $V$ is the set of $|V|=N$ nodes and each node corresponds to a deployed sensor, {\color{black}and} $E$ is the set of $|E|=M$ edges. 
The reachability between nodes, expressed as an adjacent matrix $\b{A}\in\bb{R}^{N\times N}$, could be obtained based on the pairwise road network distances between nodes.
\end{definition}
\begin{definition}
\textbf{Traffic Signal.} 
The traffic signal $\b{X}_t\in\bb{R}^{N\times C}$ denotes the observation of all sensors on the traffic network $G$ at time step $t$, where {\color{black}$C$ is the number of features collected by sensors}.
\end{definition}
\begin{definition}
\textbf{Traffic Forecasting.}
Given historical traffic signals ${\color{black} \cal{X}}={\color{black}[}\mathbf{X}_{t-T_h+1},\cdots, \mathbf{X}_{t-1},\mathbf{X}_{t}{\color{black}]}\in\bb{R}^{T_h\times N\times C}$ from the passed $T_h$ time steps, traffic forecasting aims to predict the future traffic signals ${\color{black} \cal{Y}}={\color{black}[}\mathbf{X}_{t+1},\mathbf{X}_{t+2},\cdots,\mathbf{X}_{t+T_f}{\color{black}]}$ of the $T_f$ nearest future time steps.
\end{definition}

\begin{figure*}[ht]
  \centering
  \includegraphics[width=0.93\linewidth]{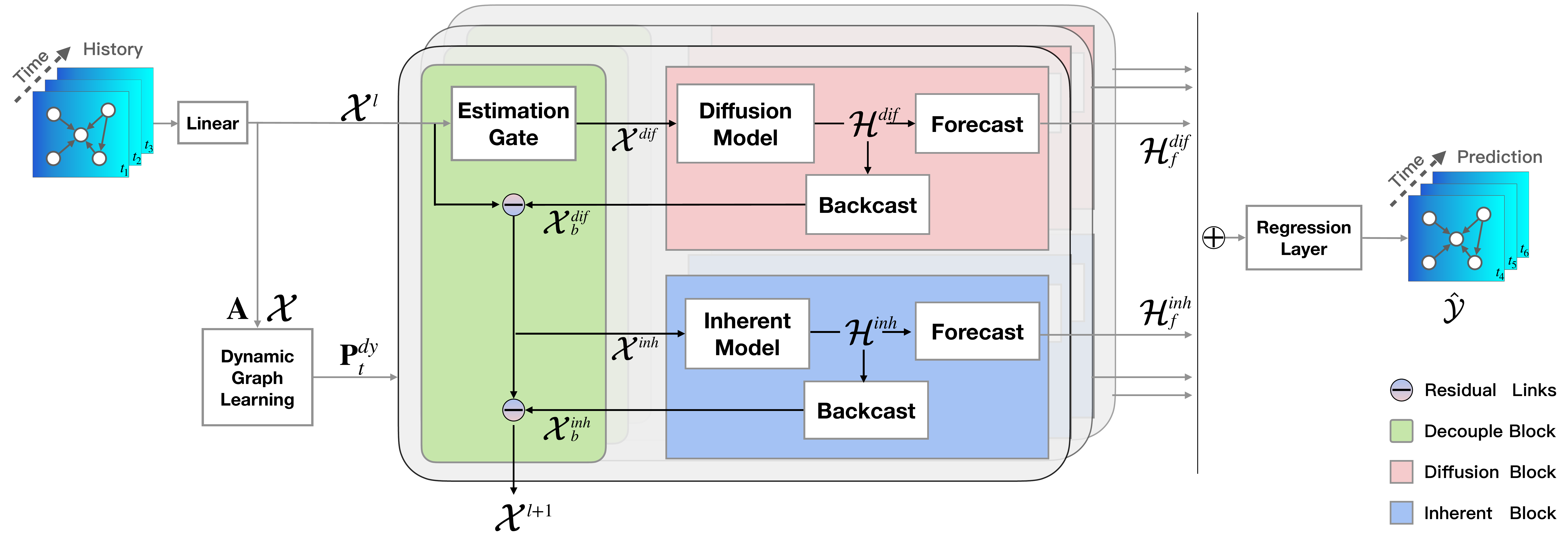}
  \caption{
The overall architecture of {\color{black}the proposed D$^2$STGNN.
The decouple block~(green) decomposes each time series in traffic signals into two hidden time series, which are subsequently handled by the diffusion block~(pink) and inherent block~(blue). 
Moreover, the dynamic graph learning module generates dynamic spatial dependency for the diffusion model.}
}

  \label{D2GNN}
\end{figure*}
\section{The Decoupled Framework}
\label{Section4}
The Decoupled Spatial-Temporal Framework~{\color{black}(DSTF)} that we propose is illustrated in Figure \ref{D2GNN}.
{\color{black}
Raw traffic signals are firstly transformed from the original space $\mathbb{R}^{T_h\times N\times C}$ to the latent space $\mathbb{R}^{T_h\times N\times d}$ by a linear layer.
For simplicity, we use $\mathcal{X}\in\mathbb{R}^{T_h\times N\times d}$ in the following as default.}
{\color{black} DSTF contains multiple decoupled spatial-temporal layers.
Given traffic signals $\mathcal{X}\in\mathbb{R}^{T_h\times N \times d}$,
the decoupled spatial-temporal layer aims at decomposing them into two hidden signals: $\mathcal{X}=\mathcal{X}^{dif}+\mathcal{X}^{inh}$, 
where $\mathcal{X}^{dif}$ and $\mathcal{X}^{inh}$ denote the diffusion signals and the inherent signals, respectively. 
However, it is a challenging task to separate them since we do not have prior knowledge.
}
{\color{black}To this end}, we propose a residual decomposition mechanism and an {\color{black}estimation} gate {\color{black}in the decouple block~(the green block) to decompose the spatial-temporal signals in a data-driven manner}.

\subsection{Residual Decomposition Mechanism}
\label{sec_res}
We first design the residual decomposition mechanism{\color{black}, which} decomposes the traffic signals {\color{black} by removing the part that has been learned by the diffusion model or inherent model in an information reconstruction fashion.}
As shown in Figure \ref{D2GNN}, {\color{black}except for the decouple block~(green), the decoupled spatial-temporal layer contains} a {\color{black}diffusion block~(pink)} and an {\color{black}inherent block~(blue)}, each with three components:
a primary model that learns knowledge from the input data {\color{black}$\mathcal{X}^*\in\bb{R}^{T_h\times N\times d}$} and generates hidden states ${\color{black}\cal{H}^*}{\color{black}\in\mathbb{R}^{T_h\times N\times d}}$, a forecast branch that generates the module's forecast hidden state ${\color{black} \cal{H}_f^*}$, a backcast branch that generates the best estimate of the module's input signal ${\color{black}\cal{X}_b^*}{\color{black}\in\mathbb{R}^{T_h\times N\times d}}$. The star indicates that this applies to both the {\color{black}diffusion and the inherent blocks}.

{\color{black}The backcast branch is crucial for the decoupling, since it reconstructs the learned knowledge,} \ie the portion of input signals that the current model can approximate well.
{\color{black}
Subsequently, residual links are designed to remove the signals that can be approximated well from the input signals and retain the signals that are not well decomposed.
Therefore, after the first~(upper) residual link, we get the input of the {\color{black}inherent} block, \ie the {\color{black}inherent} signals:
\begin{equation}
   \mathcal{X}^{\mathit{inh}}=\mathcal{X}^{l}-\mathcal{X}_b^{dif}=\mathcal{X}^{l}-\sigma(\cal{H}^{dif}\b{W}_{b}^{dif})
   \label{backcast:tem}
\end{equation}
where $\mathcal{X}^{l}$ is the input of $(l)$-th layer, and $\mathcal{X}^{0}=\mathcal{X}$.}
Superscripts {\color{black}$\mathit{dif}$ and $\mathit{inh}$} indicate {\color{black}diffusion and inherent} information, respectively.
We use non-linear fully connected networks to implement the backcast branch. $\b{W}_{b}^{\color{black}{dif}}$ is the network parameters, and the $\sigma$ is the ReLU~\cite{ReLU} activation function. 
{\color{black}
Similarly, we conduct the second~(lower) residual link after the {\color{black}inherent} block:
\begin{equation}
   \mathcal{X}^{\mathit{l+1}}=\mathcal{X}^{inh}-\mathcal{X}_b^{inh}=\mathcal{X}^{inh}-\sigma(\cal{H}^{inh}\b{W}_{b}^{inh})
   \label{backcast:next}
\end{equation}
where $\mathcal{X}^{l+1}$ retains the residual signals that can not be decomposed in the $l$-th layer. 
Similar to other deep learning methods, we stack multiple decoupled spatial-temporal layers to enhance the model's capabilities, as shown in Figure \ref{D2GNN}.}
The spatial-temporal signal can be decoupled if we design proper models for {\color{black}diffusion and inherent} signals according to their own particular characteristics, and each model can focus on its specific signals.
\subsection{Estimation Gate}
\label{sec_est}
Although the residual decomposition can decouple {\color{black}traffic} signals in a data-driven {\color{black}manner}, the {\color{black}first} model~({\color{black}the diffusion model in Figure \ref{D2GNN}}) in each layer still faces a challenge {\color{black}that may fail the decoupling process}: it takes original traffic data as input, but it needs to learn only the {\color{black}specific part of} signals in it.
To address this problem, the {\color{black} estimation} gate is designed to reduce the burden of the first model by roughly estimating the proportion of the {\color{black}two hidden time series}.
At its core, the {\color{black}estimation gate learns a gate value automatically} in ${\color{black}(}0,1{\color{black})}$ based on the current node and current time {\color{black}embeddings}.
{\color{black}Firstly,} to take into account real-world periodicities, we utilize two time slot embedding matrices: $\b{T}^D\in\bb{R}^{N_D\times d}$ and $\b{T}^W\in\bb{R}^{N_W\times d}$, where $N_D$ is the number of time slots of the day {\color{black}(determined by the sampling frequency of sensors)} and $N_W=7$ is the number of days in a week. The embeddings of time slots are thus shared among slots for the same time of the day and the same day of the week.
{\color{black} Secondly}, we use two matrices for node embeddings, the source node embedding: $\b{E}^u\in\bb{R}^{N\times d}$ is used when a node passes messages to neighboring nodes, and the target node embedding $\b{E}^d\in\bb{R}^{N\times d}$ is used when a node aggregates information from neighboring nodes.
{\color{black}Kindly note that the node embeddings and time slot embeddings are randomly initialized with learnable parameters.}
Then, given {\color{black}the} historical traffic data {\color{black}$\cal{X}$ and embeddings of time slots and nodes}, the {\color{black}estimation} gate can be written as:
\begin{equation}
\begin{aligned}
     \b{\Lambda}_{t, i}=\mathit{Sigmoid}(& \sigma((\b{T}_t^D \parallel \b{T}_t^W \parallel \b{E}_i^{u} \parallel \b{E}_i^{d})\b{W}_1)\b{W}_2)\\ 
     & {\color{black}\mathcal{X}^{\mathit{dif}}}=\Lambda\odot{\color{black}\mathcal{X}^{l}}
\end{aligned}
\label{x_spa}
\end{equation}
{\color{black}where} $\b{\Lambda}\in\bb{R}^{T_h\times N\times 1}$, and  $\b{\Lambda}_{t, i}\in (0,1)$ estimates the proportion of {\color{black}the diffusion signal in the time series in traffic data} at time slot $t$ of node $i$. 
The symbol {\color{black} $\odot$ denotes the element-wise product that broadcasts to all the channels of $\cal{X}\in\mathbb{R}^{T_h\times N\times d}$.}
$\b{W}_1\in\bb{R}^{4d\times d}$ and $\b{W}_2\in\bb{R}^{d\times 1}$ are learnable parameters, and $\sigma$ is a non-linear activation function, such as ReLU~\cite{ReLU}.

{\color{black} In addition, although we use the example of Figure \ref{D2GNN}, where the diffusion block precedes the inherent block, they are in principle interchangeable since there is no significant difference in which signal is decomposed first.
We conduct experiments in Section \ref{sec_abs} to verify that there is no significant difference in performance.
In this paper, we'll still keep the diffusion-first style.
Besides, we omit the superscript ${l}$ for each symbol of the decouple block except the input signal $\mathcal{X}^{l}$ and residual signal $\mathcal{X}^{l+1}$ for simplicity.
}

{\color{black}
In summary, this section proposes a novel framework DSTF, where each time series in traffic data is decoupled to the diffusion signals and the inherent signals in a data-driven manner.
Kindly note that other components, \ie dynamic graph learning, diffusion model, and inherent model, remain abstract and can be designed independently in the framework according to the characteristics of diffusion and inherent signals.
In the next section, we give an instantiation of DSTF by carefully designing these components.
}
\section{Decoupled Dynamic ST-GNN}
\label{Section5}
By decomposing {\color{black}diffusion and inherent} signals, the framework enables the {\color{black}subsequent} models to focus each on what they do best.
Here, we propose our Decoupled Dynamic Spatial-Temporal Graph Neural Network~(D$^2$STGNN) {\color{black}as an instantiation of the proposed framework}.
We cover the details of the {\color{black}diffusion and inherent blocks as well as the dynamic graph learning} shown in Figure \ref{D2GNN}.

\subsection{{\color{black}Diffusion Model: Spatial-Temporal} Localized {\color{black}Convolutional} Layer}
\begin{figure}[h]
  \centering
  \setlength{\abovecaptionskip}{0.2cm}
  \includegraphics[width=0.9\linewidth]{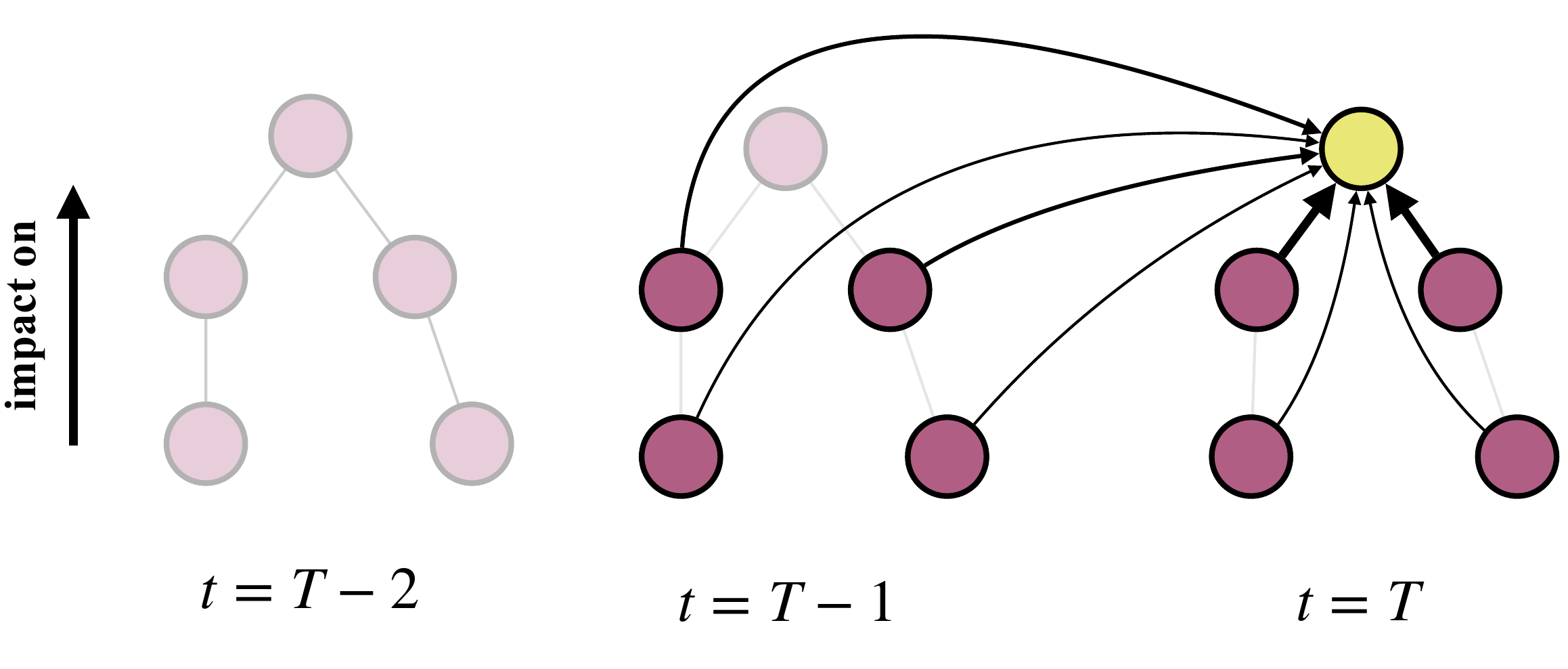}
  \caption{An example of spatial-temporal locality {\color{black}where $k_s=k_t=2$}. Only recent traffic signals from neighboring nodes can diffuse to a target node.}
  \label{locality}
\end{figure}
The {\color{black}diffusion} model aims to model {\color{black}the diffusion process between nodes, where the future diffusion signals of a target node depend on the recent values of neighboring nodes}, \ie the setting exhibits spatial-temporal locality.
Specifically, we assume that only {\color{black}the traffic signals of} $k_s$ order neighboring nodes {\color{black}from the past $k_t$ time steps} can affect a target node.
Considering the {\color{black} speed of vehicles} and the sampling frequency of sensors, the typical values of $k_s$ and $k_t$ are usually 2 or 3.
An example is shown in Figure \ref{locality}, where $k_s=k_t=2$.
In order to capture {\color{black}such the diffusion process, we design a spatial-temporal localized {\color{black}convolutional} layer.}

{\color{black}Firstly, we define} a spatial-temporal localized transition matrix:
{
\begin{equation}
    (\mathbf{P}^{\mathit{local}})^k=\underbrace{(\mathbf{P}^k\odot(1-\mathbf{I}_N))\parallel \cdots \parallel (\mathbf{P}^k\odot(1-\mathbf{I}_N))}_{k_t}
\label{P_local}
\end{equation}}where $\b{P}^k{\color{black}\in\mathbb{R}^{N\times N}}$ is a $k$ order transition matrix, and $k=1,\cdots,k_s$.
{\color{black}
Given the road network adjacency matrix $\mathbf{A}\in\mathbb{R}^{N\times N}$,} there are two directions of information diffusion: the forward transition $\b{P}_f=\b{A}/rowsum(\b{A})$, the backward transition $\b{P}_b=\b{A}^T/rowsum(\mathbf{A}^T)$.
Therefore, $(\mathbf{P}^{\mathit{local}})^k\in\bb{R}^{N\times k_tN}$, and $(\mathbf{P}^{\mathit{local}})^k[i,j]$ describes the influence of node $j$ on node $i$ in a localized spatial-temporal range.
Note that $\b{P}^{\mathit{local}}[i,i+k'N]$ ($k'=0,1,\cdots,k_t-1$) {\color{black} are masked to} zeros since they describe the inherent patterns of the target nodes themselves, which will be learned by the {\color{black}inherent} model.
For simplicity, we abbreviate $(\b{P}^{\mathit{local}})^{k}$ as {\color{black}$(\b{P}^{lc})^k$.
Secondly, c}orresponding to the Eq.~\ref{P_local}, there is a localized feature matrix {\color{black}$\b{X}^{lc}_t\in\bb{R}^{k_tN\times d}$  formed as}:
{\begin{equation}
    \mathbf{X}^{lc}_t=[\underbrace{\sigma(\mathbf{X}_{t-k_t+1}^{{\color{black}\mathit{dif}}}\mathbf{W}_{k_t-1})^T\parallel\cdots\parallel\sigma(\mathbf{X}_t^{{\color{black}\mathit{dif}}}\mathbf{W}_{0})^T}_{k_t}]^T
    \label{local_feat}
\end{equation}}where $\b{W}_k$ is the learnable parameter and $\sigma$ is the ReLU activation function. The non-linear transformation used here aims to strengthen the expressive power of the model.

{\color{black}Therefore,} based on the {\color{black} transition matrix $(\mathbf{P}^{lc})^k$ and feature matrix $\mathbf{X}^{lc}_t$} mentioned above, we define our spatial-temporal localized graph convolution operator with spatial kernels size $k_s$ as:
\begin{equation}
    \mathbf{H}_{t}^{{\color{black}\mathit{dif}}}=\sum_{k=1}^{k_s}(\mathbf{P}^{lc})^k\mathbf{X}^{lc}_t\mathbf{W}_k
    \label{ks}
\end{equation}
where $\b{H}_t^{\color{black}dif}\in\bb{R}^{N\times d}$ is the output of the localized graph convolution operator at time step $t$, which considers spatial information from the $k_s$ order neighbors.
Next, $\b{W}_k$ is the graph convolution parameters of $k$-th order, and $\b{H}_t^{\color{black}dif}$ is the hidden state of subsequent time slots that can be used to predict the {\color{black}diffusion part}.

In addition to the {\color{black}road network-based transition matrices $\mathbf{P}_b$ and $\mathbf{P}_f$}, we also utilize a self-adaptive {\color{black}transition} matrix~\cite{GWNet}.
{\color{black}
Different from the transition matrices $\mathbf{P}_b$ and $\mathbf{P}_f$, which are handcrafted by prior human knowledge, the self-adaptive transition matrix is optimized by two randomly initialized node embedding dictionaries with learnable parameters $\b{E}^u\in\mathbb{R}^{N\times d}$ and $\b{E}^d\in\mathbb{R}^{N\times d}$:
}
\begin{equation}
    \mathbf{P}_{\mathit{apt}}=\mathit{Softmax}(\sigma(\mathbf{E}^{d}(\mathbf{E}^{u})^T)).
    \label{apt}
\end{equation}
{\color{black}
Note that the $\mathbf{P}_{\mathit{apt}}\in\mathbb{R}^{N\times N}$ is normalized by the $\mathit{Softmax}$ function. 
Therefore, it describes the diffusion process that is similar to transition matrix $\mathbf{P}_b$ and $\mathbf{P}_f$.
Indeed, the matrix $\mathbf{P}_{\mathit{apt}}$ can serve as supplement to of the hidden diffusion process that is missed in the road network-based transition matrices $\mathbf{P}_b$ and $\mathbf{P}_f$.
}

{\color{black}
Given the three transition matrices $\mathbf{P}_f$, $\mathbf{P}_b$, and $\mathbf{P}_{apt}$, we can get their corresponding spatial-temporal localized transition matrix in Eq. \ref{P_local}, $(\mathbf{P}^{lc}_f)^k$, $(\mathbf{P}^{lc}_b)^k$, and $(\mathbf{P}^{lc}_{apt})^k$.
}
We can now present our localized {\color{black}convolutional} layer based on the operation in Eq. \ref{ks} as follows:
\begin{equation}
\resizebox{.92\hsize}{!}{$\mathbf{H}_{t}^{\mathit{dif}}=\sum_{k=1}^{k_s}  \left [(\mathbf{P}^{lc}_f)^k\mathbf{X}_t^{lc}\mathbf{W}_{k1}+(\mathbf{P}^{lc}_b)^k\mathbf{X}_t^{lc}\mathbf{W}_{k2}+{\color{black}(\mathbf{P}^{lc}_{apt})^{k}}\mathbf{X}_t^{lc}\mathbf{W}_{k3}\right ].$}
    \label{stconv}
\end{equation}

In summary, given temporal kernel size $k_t$ and spatial kernel size $k_s$ as well as input ${\color{black}\cal{X}}^{\color{black}dif}\in\bb{R}^{T_h\times N\times d}$, the localized {\color{black}convolutional} layer generates a hidden state sequence ${\color{black}\cal{H}^{\color{black}dif}}$ by synchronously modeling the spatial-temporal correlations in {\color{black}each time step $t$}:
\begin{equation}
    {\color{black}\mathcal{H}^{\mathit{\color{black}dif}}}=\mathbf{\Theta}_{*G}(\mathcal{X}^{\color{black}dif})=[\cdots, \mathbf{H}_{T-2}^{\mathit{\color{black}dif}},\mathbf{H}_{T-1}^{\mathit{\color{black}dif}},\mathbf{H}_{T}^{\mathit{\color{black}dif}}]
    \label{eq:hspa}
\end{equation}
where $\b{\Theta}$ denotes all the parameters mentioned in Eqs.~\ref{local_feat}, \ref{apt}, and \ref{stconv}.
{\color{black}$\*G$ denotes the spatial-temporal localized convolution in Eq.~\ref{stconv}.}
The output hidden state sequence ${\color{black}\cal{H}^{\mathit{\color{black}dif}}}$ is further used to generate two outputs, \ie the backcast and forecast output.

\noindent\textbf{Forecast Branch}: The last hidden state $\b{H}_T^{\mathit{\color{black}dif}}$ can be used to forecast the value of the next step. 
In order to forecast the hidden state in multi-step forecasting task, we follow an auto-regressive procedure to generate ${\color{black}\cal{H}^{\mathit{\color{black}dif}}_f} = [\b{H}_{T+1}^{\mathit{\color{black}dif}}, \b{H}_{T+2}^{\mathit{\color{black}dif}}, \dots, \b{H}_{T+T_f}^{\mathit{\color{black}dif}}]$, each of which is used by a non-linear regression neural network to predict the particular values that we are interested in. 

\noindent\textbf{Backcast Branch}: 
{\color{black}
As discussed in Section \ref{sec_res}, we use non-linear fully connected networks to implement the backcast branch, and generate $\mathcal{X}_b^{\color{black}dif}=\sigma(\cal{H}^{\color{black}dif}\b{W}_{b}^{\color{black}dif})$, \ie the learned {\color{black}diffusion} part, which subsequently is removed from the original signals by the residual link in Eq. \ref{backcast:tem} to achieve decomposition.}

\subsection{{\color{black}Inherent} Model: Local and Global Dependency}
\begin{figure}[h]
  \setlength{\abovecaptionskip}{0.2cm}
  \centering
  \includegraphics[width=0.8\linewidth]{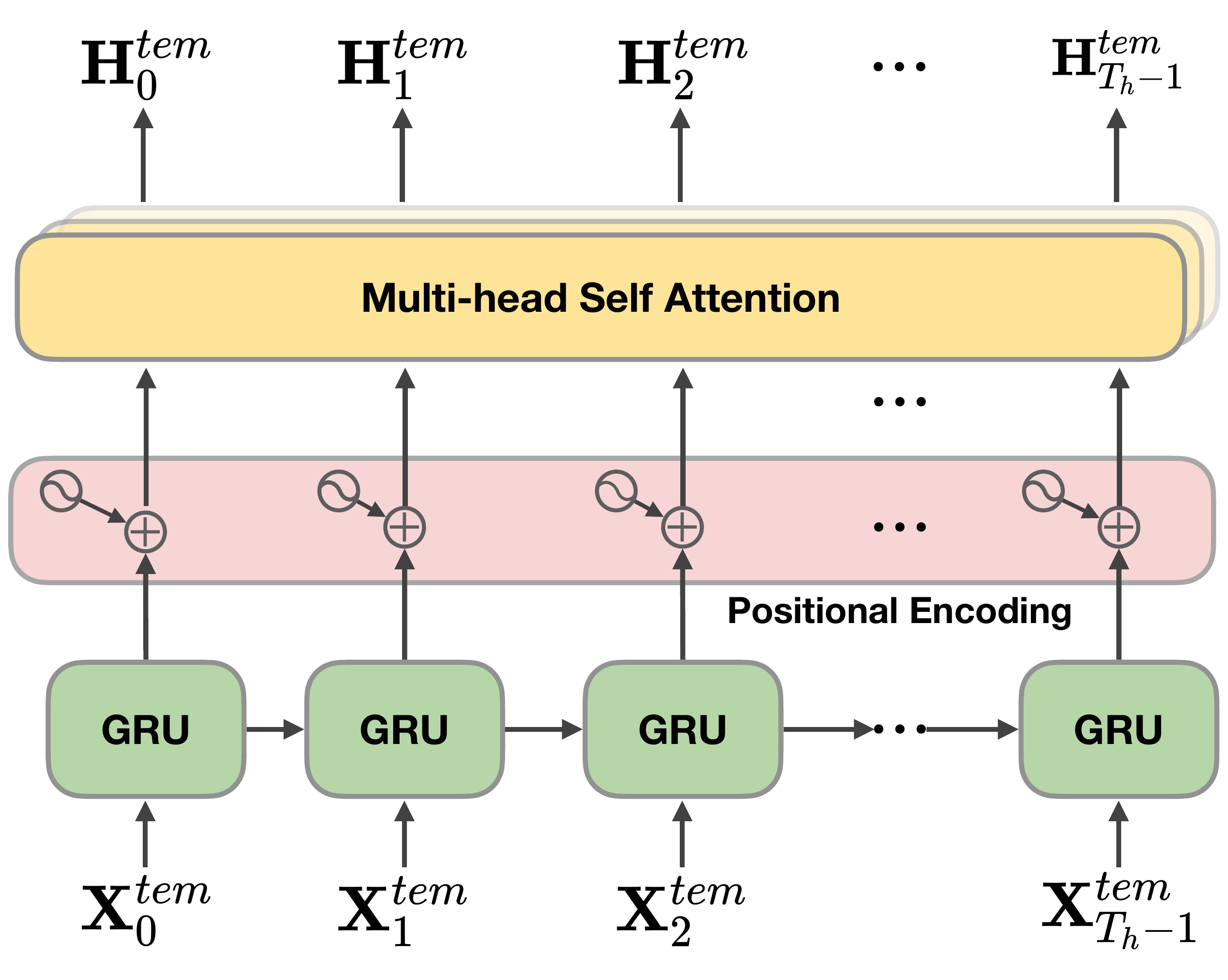}
  \caption{The {\color{black}inherent} model. Short-term dependencies are captured by the GRU, while long-term dependencies are captured by the multi-head self-attention layer.}
  \label{temporal}
\end{figure}

The {\color{black}inherent model is designed to model the \textit{hidden inherent time series} in the original signals of each node, \ie the $\mathcal{X}^{inh}$}. 
Dependencies in time series are often divided into local and global dependencies, a.k.a. short- and long-term dependencies~\cite{2020STGNN, 2020NBeats, TrendPeriodic}.
Previous studies have shown that Gated Recurrent Units~\cite{2014GRU}~(GRUs) are good at capturing short-term dependencies, while a self-attention layer does better at handling long-term dependencies~\cite{2020STGNN}.
We utilize GRU~\cite{2014GRU} and a multi-head self-attention layer~\cite{2017Transformer} jointly to capture temporal patterns comprehensively.
A diagram of the {\color{black}inherent} model is shown in Figure \ref{temporal}.

GRU can recurrently preserve the hidden state of history data and control the information that flows to the next time step.
Given the input signal of {\color{black}inherent block} $\b{X}^{\mathit{\color{black}inh}}_t\in\bb{R}^{N\times d}$ at time step $t$ , for each node $i$, we use the following GRU operation:
\begin{equation}
    \begin{aligned}
         \mathbf{z}_t&=\sigma(\mathbf{W}_z\mathbf{X}^{\mathit{\color{black}inh}}_t[i,:]+\mathbf{U}_z\mathbf{H}^{\mathit{\color{black}inh}}_{t-1}[i,:]+\mathbf{b}_z)\\[1.2mm]
         \mathbf{r}_t&=\sigma(\mathbf{W}_r\mathbf{X}^{\mathit{\color{black}inh}}_t[i,:]+\mathbf{U}_r\mathbf{H}^{\mathit{\color{black}inh}}_{t-1}[i,:]+\mathbf{b}_r)\\[1.2mm]
         \hat{\mathbf{H}}^{\mathit{\color{black}inh}}_{t}[i,:]&=tanh(\mathbf{W}_h\mathbf{X}^{\mathit{\color{black}inh}}_t[i, :] +\mathbf{r}_t\odot(\mathbf{U}_h\mathbf{H}^{\mathit{\color{black}inh}}_{t-1}[i, :]+\mathbf{b}_h))\\[1.2mm]
         \widetilde{\mathbf{H}}^{\mathit{\color{black}inh}}_{t}[i,:]&=(1-\mathbf{z}_t)\odot\hat{\mathbf{H}}^{\mathit{\color{black}inh}}_{t-1}[i,:]+\mathbf{z}_t\odot\hat{\mathbf{H}}^{\mathit{\color{black}inh}}_t[i,:]
    \end{aligned}
\end{equation}
where $\widetilde{\mathbf{H}}^{\mathit{\color{black}inh}}_{t}[i, :]$ is the updated hidden state of node $i$ at time step $t$, $\odot$ denotes the element-wise product, and $\b{W}_z$, $\b{W}_r$, $\b{W}_h$, $\b{U}_z$, $\b{U}_r$, and $\b{U}_h$ are the learnable parameters of GRU. 

The GRU can capture short-term sequential information well. 
However, capturing only local information is insufficient because traffic forecasting is also affected by longer-term dependencies~\cite{2020STGNN}. 
Hence we introduce a multi-head self-attention layer to capture global dependencies on the top of the GRU.
{\color{black}Given the output of the GRU, {\color{black} $\widetilde{\cal{H}}^{inh}\in\bb{R}^{T_h\times N\times d}$}, the multi-head self-attention layer performs pair-wisely dot product attention on the time dimension for each node, \ie the product is calculated between any two signals of different time slots.}
Therefore, the receptive field is theoretically infinite, which is beneficial to capturing global dependencies.
Specifically, given attention head $s$, the learnable project matrices $\mathbf{W}_s^{Q},\mathbf{W}_s^{K},\mathbf{W}_s^{V}\in\bb{R}^{d\times d}$, and the output matrix $\b{W}^O$, the attention function of node $i$ can be written as:
\begin{equation}
 \begin{aligned}
    {\color{black}\cal{H}^{\mathit{\color{black}inh}}}[:,i,:]&=\mathit{Multihead}(\mathbf{H}^{v_i})\\
    &=\mathit{Concat}(\mathit{head}_1,\cdots,\mathit{head}_S)\mathbf{W}^O\\
    {\rm where}\ \mathit{head}_s&=\ \mathit{Attention}_s(\mathbf{H}^{v_i}) \\
    &=\ \mathit{softmax}(\frac{\mathbf{H}^{v_i}\mathbf{W}_s^Q(\mathbf{H}^{v_i}\mathbf{W}_s^{K})^T}{\sqrt{d}}\mathbf{H}^{v_i}\mathbf{W}_s^V)
  \end{aligned}
  \label{mhsa}
\end{equation}
where $\b{H}^{v_i}\in\bb{R}^{T\times d}$ is the feature of node $v_i$ in all time slots. 
All the nodes are calculated individually in parallel with the help of the GPU. 
Hence, we can get the hidden state of {\color{black}inherent} model ${\color{black}\cal{H}^{\mathit{\color{black}inh}}}\in\bb{R}^{T\times N\times d}$.
Although the self-attention layer has an infinite receptive field, it ignores relative positions in the sequence.
To take into account the position, we apply positional encoding~\cite{2017Transformer} between GRU and multi-head self-attention layer as follows:
\begin{equation}
\begin{aligned}
     &{\color{black}\widetilde{\mathbf{H}}^{\mathit{\color{black}inh}}_{t}[i,:]=\widetilde{\mathbf{H}}^{\mathit{\color{black}inh}}_{t}[i,:]+\mathbf{e}_t} \\
     & \mathbf{e}_{t, i}=   \left\{
                    \begin{aligned} 
                            sin(t/10000^{2i/d}),\quad &\mathit{if}\ i = 0, 2, 4... \\
                            cos(t/10000^{2i/d}),\quad &\mathit{otherwise}
                    \end{aligned} 
                    \right.
\end{aligned}
\end{equation}
where $\b{e}_t\in\bb{R}^{d}$ is the positional embedding of time step $t$. Note that the positional encoding is not trainable.

\noindent\textbf{Forecast Branch}: Here, we also adopt auto-regression to generate the future hidden state ${\color{black}\cal{H}^{\mathit{\color{black}inh}}_f} = [\b{H}_{T+1}^{\mathit{\color{black}inh}},  \b{H}_{T+2}^{\mathit{\color{black}inh}}, \dots, \b{H}_{T+T_f}^{\mathit{\color{black}inh}}]$. 
Specifically, we adopt a simple sliding auto-regression, rather than the {\color{black}commonly used} encoder-decoder architecture~\cite{2014Seq2Seq, 2017Transformer} because we do not have the ground truth of {\color{black}\textit{hidden inherent time series}}, which are crucial when having to train a decoder.

\noindent\textbf{Backcast Branch}:
{\color{black}
Same as the {\color{black}diffusion} block, we use non-linear  fully connected networks to implement the backcast branch and generate $\mathcal{X}_b^{\color{black}inh}=\sigma(\cal{H}^{\color{black}inh}\b{W}_{b}^{\color{black}inh})$, \ie the learned {\color{black}inherent} part, which is subsequently used in Eq. \ref{backcast:next}.}

\subsection{Dynamic Graph Learning}
{\color{black} In this subsection}, we design a dynamic graph learning model to {\color{black}capture the dynamics of spatial dependency, as discussed in Section \ref{Section1}}.
{\color{black} The intensity of traffic diffusion between two connected nodes changes dynamically over time in the real world.
An example is shown in Figure \ref{Intro1}(c), where the influence between nodes is different between 8:00 am and 10:00 am.}
Therefore, it is crucial to model dynamic transition {\color{black} matrices $\b{P}^{\mathit{dy}}_b$ and $\b{P}^{\mathit{dy}}_f$ to enhance the static ones $\b{P}_b$ and $\b{P}_f$ by replacing them in Eqs. \ref{stconv} and \ref{P_local}.} 

The core of modeling {\color{black} $\b{P}^{\mathit{dy}}_b$ and $\b{P}^{\mathit{dy}}_f$} is to ensure that the static, dynamic, and time information in traffic data is encoded comprehensively.
{\color{black}
For a given time step $t$, we take the historical observation as the dynamic feature.
For example, {\color{black}given historical data $\mathcal{X}\in\bb{R}^{T_h\times N\times d}$},
the dynamic information of channel $c$ can be formulated as  $\b{X}_c=\cal{X}[:, :, c]^T\in\bb{R}^{N\times T}$, where $c = 1, ..., d$.
In addition, we consider {\color{black}the time embeddings $\mathbf{T}^D_t\in\bb{R}^{d}$ and $\mathbf{T}^W_t \in\bb{R}^{d}$}, which are the embeddings used in the estimation gate in Section \ref{sec_est}.}
Employing also two static node embedding matrices $\b{E}^u\in\bb{R}^{N\times d}$ and $\b{E}^d\in\bb{R}^{N\times d}$, we first obtain two dynamic feature matrices:
\vspace{-0.1cm}
\begin{equation}
    \begin{split}
      & \b{DF}^u_{t}=\mathit{Concat}[\text{FC}(\mathop{\parallel}\limits_{c=1}^{C}\b{X}_c),{\color{black}\b{T}^D_t,\b{T}^W_t},\b{E}^u]\\
      & \b{DF}^d_{t}=\mathit{Concat}[\text{FC}(\mathop{\parallel}\limits_{c=1}^{C}\b{X}_c),{\color{black}\b{T}^D_t,\b{T}^W_t},\b{E}^d].
    \end{split}
\end{equation}
Here, $\b{DF}_t^u\in\bb{R}^{N\times 4d}$, and $\b{DF}_t^d\in\bb{R}^{N\times 4d}$.
And $\text{FC}(\cdot)$ is a non-linear two-layer fully connected network that extracts features and transforms the dimensionality from $N \times dT$ to $N \times d$. 
Further, $\mathit{Concat}(\cdot)$ denotes broadcast concatenation.
Then we use the attention mechanism to calculate the pair-wise mask to get dynamic graphs:
\vspace{-0.1cm}
\begin{equation}
    \begin{split}
     &\b{P}^{\mathit{dy}}_{f, t}=\b{P}_f\odot \mathit{Softmax}(\frac{(\b{DF}_t^u\b{W}^Q)(\b{DF}_t^u\b{W}^K)^T}{\sqrt{d}})\\
     &\b{P}^{\mathit{dy}}_{b, t}=\b{P}_b\odot \mathit{Softmax}(\frac{(\b{DF}_t^d\b{W}^Q)(\b{DF}_t^d\b{W}^K)^T}{\sqrt{d}}).\\
\end{split}
\label{dy}
\end{equation}
{\color{black}$\mathbf{W}_Q$ and $\mathbf{W}_K$ are the parameters of self-attention mechanism~\cite{2017Transformer}.}
Matrices $\b{P}^{\mathit{dy}}_{f, t}$ and $\b{P}^{\mathit{dy}}_{f, t}\in\bb{R}^{N\times N}$ can replace the transition matrices in Eqs.~\ref{stconv} and \ref{P_local} to enhance the model, thus completing the proposed D$^2$STGNN model. In practice, the calculation of the adjacency matrix is expensive, so to reduce the computational cost, we assume that given a limited time range $T_h$, $\b{P}^{\mathit{dy}}$ is static, \ie $\b{P}^{\mathit{dy}}_{t-T_h:t}=\b{P}^{\mathit{dy}}_{t}$.

\subsection{Output and Training Strategy}
{\color{black}Assuming we stack $L$ decoupled spatial-temporal layers, we include the output hidden states in the forecast branches $\mathcal{H}_f^{{\color{black}dif}, l}=[\mathbf{H}_{T+1}^{{\color{black}dif}, l}, \cdots]$ and $\mathcal{H}_f^{{\color{black}inh}, l}=[\mathbf{H}_{T+1}^{{\color{black}inh}, l}, \cdots]$ of {\color{black}the diffusion and the inherent} blocks of the $l$-th layer to generate our final forecasting}:
\begin{equation}
    \begin{split}
      \cal{H} &= \cal{H}_f^{\mathit{{\color{black}dif}}} + \cal{H}_f^{\mathit{{\color{black}inh}}}=
    \sum_{l=0}^{L-1}\mathcal{H}_f^{{\color{black}dif}, l} + \sum_{l=0}^{L-1}\mathcal{H}_f^{{\color{black}inh}, l}\\[-2mm]
      & = [\sum_{l=0}^{L-1}(\b{H}^{\mathit{{\color{black}dif}, l}}_{T+1}+\b{H}^{\mathit{{\color{black}inh}, l}}_{T+1}), \sum_{l=0}^{L-1}(\b{H}^{\mathit{{\color{black}dif}, l}}_{T+2}+\b{H}^{\mathit{{\color{black}inh}, l}}_{T+2}), \cdots].\\
    \end{split}
\end{equation}
{\color{black}Then we adopt a two-layer fully connected network as our regression layer and apply it to $\cal{H}$ to generate the final predictions.
The outputs of the regression layer at each time step are concatenated to form the final output:}
$\hat{\mathcal{Y}}\in\bb{R}^{T_f\times N \times C_{\mathit{out}}}$. 
Given the ground truth \noindent$\cal{Y}\in\bb{R}^{T_f\times N \times C_{\mathit{out}}}$, {\color{black} we optimize our model using MAE loss}:
\vspace{-0.2cm}
\begin{equation}
    \cal{L}(\hat{\cal{Y}}, \cal{Y};\b{\Theta})=\frac{1}{T_fNC_{\mathit{out}}}
    \sum_{i=1}^{T_f}\sum_{j=1}^{N}\sum_{k=1}^{C_{\mathit{out}}}|\hat{\cal{Y}}_{ijk} - \cal{Y}_{ijk}|
    \label{loss}
\end{equation}
where $N$ is the number of nodes, $T_f$ is the number of forecasting steps, and $C_{\mathit{out}}$ is the dimensionality of the output.
Following existing studies \cite{2021DGCRN, 2020MTGNN}, we employ curriculum learning, a general and effective training strategy, to train the proposed model. 
We optimize the model parameters by minimizing $\cal{L}$ via gradient descent. 
The overall learning algorithm is outlined in Algorithm \ref{alg}.

\normalem

\begin{algorithm}[tp]
\caption{The overall learning algorithm of D$^2$STGNN}
\LinesNumbered
\label{alg}
\KwIn{ 
The traffic signals over the past $T_h$ time steps $\cal{X}$,
the adjacency matrix $\b{A}$,
time embeddings $\b{T}^D$ and $\b{T}^W$,
node embeddings $\b{E}^d$ and $\b{E}^u$,
the number of layers $L$.
}
\KwOut{The prediction of traffic signals $\mathcal{Y}$ in $T_f$ future time steps.}

{
Calculate self-adaptive adjacent matrix $\b{A}_{apt}$ by Eq.~\ref{apt}.

Calculate dynamic transition matrix $\b{P}^{\mathit{dy}}_f$ and $\b{P}^{\mathit{dy}}_b$ by Eq.~\ref{dy}.

output $\leftarrow$ [ ]

$\cal{X}^0\leftarrow\cal{X}$
}

{
\setstretch{0.96}
\For {l in \textbf{range}(L) }{
Calculate $\cal{X}^{\mathit{{\color{black}dif}}}$ according to Eq.~\ref{x_spa} with time and node embeddings. {\color{black}\algorithmiccomment{{\color{black}Estimation} gate}}\\
Calculate $\cal{H}^{\mathit{{\color{black}dif}}},\cal{X}^{\mathit{{\color{black}dif}}}_b$ according to Eq.~\ref{eq:hspa} and the backcast branch. {\color{black}\algorithmiccomment{{\color{black}Diffusion block}}}\\
Calculate $\cal{X}^{\mathit{{\color{black}inh}}}$ according to Eq.~\ref{backcast:tem}.{\color{black}\algorithmiccomment{Decomposition}}\\
Calculate $\cal{H}^{\mathit{{\color{black}inh}}},\cal{X}_b^{\mathit{{\color{black}inh}}}$ according to Eq.~\ref{mhsa} and the backcast branch. {\color{black}\algorithmiccomment{{\color{black}Inherent block}}}\\
Calculate $\cal{X}^{l+1}$ according to Eq.~\ref{backcast:next}.{\color{black}\algorithmiccomment{Decomposition}}\\
Append the output of forecast branch of {\color{black}diffusion and inherent block} to the output list. 
}
}
$\cal{H}\leftarrow\text{sum(output)}$\\
$\hat{\cal{Y}}\leftarrow\text{MLP}(\cal{H})$\\
Backpropagation and update parameters according to Eq.~\ref{loss}.
\end{algorithm}

\section{Experiments}
\label{Section6}
In this section, we present experiments on four large real-world datasets to demonstrate the effectiveness of D$^2$STGNN for traffic forecasting. 
We first introduce the experimental settings, including datasets, baselines, and parameter settings.
Then, we conduct experiments to compare the performance of the D$^2$STGNN with other baselines.
Furthermore, we design more experiments to verify the superiority of our decoupling framework. 
Finally, we design {\color{black}comprehensive} ablation studies to evaluate the impact of the essential {\color{black}architectures, }components{\color{black}, and} training strategies.

\subsection{Experimental Setup}
\noindent\textbf{Datasets.} 
We conducted experiments on four commonly used real-world large-scale datasets, which have tens of thousands of time steps and hundreds of sensors.
The statistical information is summarized in Table \ref{tab:datasets}.
Two of them are traffic speed datasets, while the others are traffic flow dataset. 
Traffic speed data records the average vehicles speed (miles per hour). 
Due to the speed limit in these areas, the speed data is a float value usually less than 70.
The flow data should be an integer, up to hundreds, because it records the number of passing vehicles.
All these datasets have one feature channel~(the traffic speed or the traffic flow), \ie $C=1$.

\noindent\textbf{Construction of the traffic network.}
For traffic speed datasets, we follow the procedure of DCRNN~\cite{2017DCRNN}.
We compute the pairwise road network distances between sensors and build the adjacency matrix using thresholded Gaussian kernel~\cite{Gaussian}. 
For traffic flow datasets, we use the traffic network provided by ASTGCN~\cite{2019ASTGCN}. They remove many redundant detectors to ensure the distance between any adjacent detectors is longer than 3.5 miles to obtain a lightweight traffic network. The following gives more detailed description of the four datasets:

\begin{itemize}
    \item \textbf{METR-LA} is a public traffic speed dataset collected from loop-detectors located on the LA County road network~\cite{METR-LA}. 
    Specifically, METR-LA contains data of 207 sensors over a period of 4 months from Mar 1st 2012 to Jun 30th 2012~\cite{2017DCRNN}. 
    The traffic information is recorded at the rate of every 5 minutes, and the total number of time slices is 34,272.
    \item \textbf{PEMS-BAY} is a public traffic speed dataset collected from California Transportation Agencies (CalTrans) Performance Measurement System (PeMS)~\cite{PEMS-BAY}. 
    Specifically, PEMS-BAY contains data of 325 sensors in the Bay Area over a period of 6 months from Jan 1st 2017 to May 31th 2017~\cite{2017DCRNN}.
    The traffic information is recorded at the rate of every 5 minutes, and the total number of time slices is 52,116.
    \item 
    \color{black}{\textbf{PEMS04} is a public traffic flow dataset collected from CalTrans PeMS~\cite{PEMS-BAY}. 
    Specifically, PEMS04 contains data of 307 sensors in the District04 over a period of 2 months from Jan 1st 2018 to Feb 28th 2018~\cite{2019ASTGCN}.
    The traffic information is recorded at the rate of every 5 minutes, and the total number of time slices is 16,992.}
    \item
    \color{black}{\textbf{PEMS08} is a public traffic flow dataset collected from CalTrans PeMS~\cite{PEMS-BAY}. 
    Specifically, PEMS08 contains data of 170 sensors in the District08 over a period of 2 months from July 1st 2018 to Aug 31th 2018~\cite{2019ASTGCN}.
    The traffic information is recorded at the rate of every 5 minutes, and the total number of time slices is 17,833.}
\end{itemize}

\begin{table}
\caption{Statistics of datasets.}
\label{tab:datasets}
  \begin{tabular}{c|c|c|c|c}
    \toprule
    \textbf{Type} &\textbf{Dataset} & \textbf{\# Node} & \textbf{\# Edge} & \textbf{\# Time Step}\\
    \midrule
    \multirow{2}*{\textbf{Speed}} &
    \textbf{METR-LA} & 207 &1722 & 34272\\
    & \textbf{PEMS-BAY} & 325 &2694 & 52116\\
    
    \hline
    \multirow{2}*{\textbf{Flow}}  & 
    \textbf{PEMS04} & 307 &680 & 16992\\
    & \textbf{PEMS08} & 170 &548 & 17856\\
    \bottomrule
  \end{tabular}
\end{table}

\noindent\textbf{Baselines.} We select a wealth of baselines that have official public code, including the traditional methods and the typical deep learning methods, as well as  the very recent state-of-the-art works.
\begin{itemize}
    \item \textbf{HA}: 
    Historical Average model, which models traffic flows as a periodic process and uses weighted averages from previous periods as predictions for future periods.
    \item \textbf{VAR}: 
    Vector Auto-Regression~\cite{VAR, lutkepohl2005new} assumes that the passed time series is stationary and estimates the relationship between the time series and their lag value.~\cite{2020STGNN}
    \item \textbf{SVR}: Support Vector Regression (SVR) uses linear support vector machine for classical time series regression task.
    \item \textbf{FC-LSTM}~\cite{FC-LSTM}: Long Short-Term Memory network with fully connected hidden units is a well-known network architecture that is powerful in capturing sequential dependency.
    \item \textbf{DCRNN}~\cite{2017DCRNN}: Diffusion Convolutional Recurrent Neural Network~\cite{2017DCRNN} models the traffic flow as a diffusion process. It replaces the fully connected layer in GRU~\cite{2014GRU} by diffusion convolutional layer to form a new Diffusion Convolutional Gated Recurrent Unit~(DCGRU).
    \item \textbf{Graph WaveNet}~\cite{GWNet}: Graph WaveNet stacks Gated TCN and GCN layer by layer to jointly capture the spatial and temporal dependencies.
    \item \textbf{ASTGCN}~\cite{2019ASTGCN}: ASTGCN combines the spatial-temporal attention mechanism to capture the dynamic spatial-temporal characteristics of traffic data simultaneously.
    \item \textbf{STSGCN}~\cite{2020STSGCN}: STSGCN is proposed to effectively capture the localized spatial-temporal correlations and consider the heterogeneity in spatial-temporal data.
    \item \textbf{GMAN}~\cite{2020GMAN}: GMAN is an attention-based model which stacks spatial, temporal and transform attentions.
    \item \textbf{MTGNN}~\cite{2020MTGNN}: MTGNN extends Graph WaveNet through the mix-hop propagation layer in the spatial module, the dilated inception layer in the temporal module, and a more delicate graph learning layer.
    \item \textbf{DGCRN}~\cite{2021DGCRN}: DGCRN models the dynamic graph and designs a novel Dynamic Graph Convolutional Recurrent Module~(DGCRM) to capture the spatial-temporal pattern in a seq2seq architecture. 
\end{itemize}
We use the default settings as described in baseline papers. 
We evaluate the performances of all baselines by three commonly used metrics in traffic forecasting, including Mean Absolute Error~(MAE), Root Mean Squared Error~(RMSE) and Mean Absolute Percentage Error~(MAPE).
The formulas are as follows:
\begin{equation}
\begin{aligned}
    & \text{MAE}(x, \hat{x})=\frac{1}{|\Omega|} \sum_{i\in\Omega}|x_i - \hat{x}_i|, \\
    & \text{RMSE}(x, \hat{x})=\sqrt{\frac{1}{|\Omega|} \sum_{i\in\Omega}(x_i - \hat{x}_i)^2}, \\
    & \text{MAPE}(x, \hat{x})=\frac{1}{|\Omega|} \sum_{i\in\Omega}\frac{|x_i - \hat{x}_i|}{x_i},
\end{aligned}
\end{equation}
where MAE metric reflects the prediction accuracy~\cite{2021DGCRN}, RMSE is more sensitive to abnormal values, and MAPE can eliminate the influence of data units to some extent. $x_i$ denotes the $i$-th ground truth, $\hat{x}_i$ represents the $i$-th predicted values, and $\Omega$ is the indices of observed samples, where $|\Omega|=T_f=12$ in our experiments.

\noindent\textbf{Implementation.}
The proposed model is implemented by Pytorch 1.9.1 on NVIDIA 3090 GPU.
We use Adam as our optimizer and set the learning rate to 0.001.
The embedding size of nodes and time slots is set to 12.
The other hidden dimension $d$ in this paper is set to 32.
The spatial kernel size is set to 2, {\color{black} and the temporal kernel size is set to 3 for all datasets}.
The batch size is set to 32.
We employ early stopping to avoid overfitting.
We perform significance test (t-test with p-value $< 0.05$) over all the experimental results.
For any other more details, readers could refer to our public code repository.

\subsection{The Performance of D$^2$STGNN}
\subsubsection{Settings}
For a fair comparison, we follow the dataset division in previous works.
For METR-LA and PEMS-BAY, we use about 70\% of data for training, 20\% of data for testing, and the remaining 10\% for validation~\cite{2017DCRNN, GWNet, 2020MTGNN, 2021DGCRN}.
For PEMS04 and PEMS08, we use about 60\% of data for training, 20\% of data for testing, and the remaining 20\% for validation~\cite{2021ASTGNN, 2019ASTGCN, 2020STSGCN}.
We generate sequence samples through a sliding window with a width of 24~(2 hours), where the first 12 time steps are used as input, and the remaining 12 time {\color{black}steps} are used as ground truth.
We compare the performance of 15 minutes~(horizon 3), 30 minutes~(horizon 6), and 1 hour~(horizon 12) ahead forecasting {\color{black} on the MAE, RMSE, and MAPE metrics}.
\subsubsection{Results}
\begin{table*}[p]
\renewcommand\arraystretch{0.912}
    \centering
    \caption{
      Traffic forecasting on the METR-LA, PEMS-BAY, PEMS04,and PEMS08 datasets. Numbers marked with $^*$ indicate that the improvement is statistically significant compared with the best baseline~(t-test with p-value$<0.05$).}
    \label{tab:main}
    \begin{tabular}{ccccr|ccr|ccr}
      \toprule
      \midrule  
      \multirow{2}*{\textbf{Datasets}} &\multirow{2}*{\textbf{Methods}} & \multicolumn{3}{c}{\textbf{Horizon 3}} & \multicolumn{3}{c}{\textbf{Horizon 6}}& \multicolumn{3}{c}{\textbf{Horizon 12}}\\ 
      \cmidrule(r){3-5} \cmidrule(r){6-8} \cmidrule(r){9-11}
      &  & MAE & RMSE & MAPE & MAE & RMSE & MAPE & MAE & RMSE & MAPE\\
      \midrule
      \hline
      \multirow{14}*{\textbf{METR-LA}} 
      &HA              & 4.79  & 10.00 & 11.70\%       & 5.47  & 11.45 & 13.50\%      & 6.99  & 13.89  & 17.54\% \\ 
      &VAR             & 4.42  & 7.80  & 13.00\%       & 5.41  & 9.13  & 12.70\%      & 6.52  & 10.11 & 15.80\% \\ 
      &SVR             & 3.39  & 8.45  & 9.30\%        & 5.05  & 10.87 & 12.10\%      & 6.72  & 13.76 & 16.70\% \\ 
      &FC-LSTM         & 3.44  & 6.30  & 9.60\%        & 3.77  & 7.23  & 10.09\%      & 4.37  & 8.69  & 14.00\% \\ 
      &DCRNN           & 2.77  & 5.38  & 7.30\%        & 3.15  & 6.45  & 8.80\%       & 3.60  & 7.60  & 10.50\% \\ 
      &STGCN           & 2.88  & 5.74  & 7.62\%        & 3.47  & 7.24  & 9.57\%       & 4.59  & 9.40  & 12.70\% \\ 
      &Graph WaveNet   & 2.69  & 5.15  & 6.90\%        & 3.07  & 6.22  & 8.37\%       & 3.53  & 7.37  & 10.01\% \\
      &ASTGCN          & 4.86  & 9.27  & 9.21\%        & 5.43  & 10.61 & 10.13\%      & 6.51  & 12.52 & 11.64\% \\  
      &STSGCN          & 3.31  & 7.62  & 8.06\%        & 4.13  & 9.77  & 10.29\%      & 5.06  & 11.66 & 12.91\% \\  
      &MTGNN           & 2.69  & 5.18  & 6.88\%        & 3.05  & 6.17  & 8.19\%       & 3.49  & 7.23  & 9.87\% \\  
      &GMAN            & 2.80  & 5.55  & 7.41\%        & 3.12  & 6.49  & 8.73\%       & 3.44  & 7.35  & 10.07\% \\  
      &DGCRN           & 2.62  & 5.01  & 6.63\%        & 2.99  & 6.05  & 8.02\%       & 3.44  & 7.19  & 9.73\% \\  
    \cmidrule(r){2-11}
    &D$^2$STGNN      & \textbf{2.56}$^*$  & \textbf{4.88}$^*$  & \textbf{6.48\%}$^*$        & \textbf{2.90}$^*$  & \textbf{5.89}$^*$  & \textbf{7.78\%}$^*$      & \textbf{3.35}$^*$  & \textbf{7.03}$^*$  & \textbf{9.40\%}$^*$ \\ 
    \midrule
    \hline
    \multirow{14}*{\textbf{PEMS-BAY}} 
      &HA              & 1.89  & 4.30  & 4.16\%        & 2.50  & 5.82  & 5.62\%       & 3.31  & 7.54  & 7.65\% \\ 
      &VAR             & 1.74  & 3.16  & 3.60\%        & 2.32  & 4.25  & 5.00\%       & 2.93  & 5.44  & 6.50\% \\ 
      &SVR             & 1.85  & 3.59  & 3.80\%        & 2.48  & 5.18  & 5.50\%       & 3.28  & 7.08  & 8.00\% \\ 
      &FC-LSTM         & 2.05  & 4.19  & 4.80\%        & 2.20  & 4.55  & 5.20\%       & 2.37  & 4.96  & 5.70\% \\ 
      &DCRNN           & 1.38  & 2.95  & 2.90\%        & 1.74  & 3.97  & 3.90\%       & 2.07  & 4.74  & 4.90\% \\ 
      &STGCN           & 1.36  & 2.96  & 2.90\%        & 1.81  & 4.27  & 4.17\%       & 2.49  & 5.69  & 5.79\% \\ 
      &Graph WaveNet   & 1.30  & 2.74  & 2.73\%        & 1.63  & 3.70  & 3.67\%       & 1.95  & 4.52  & 4.63\% \\
      &ASTGCN          & 1.52  & 3.13  & 3.22\%        & 2.01  & 4.27  & 4.48\%       & 2.61  & 5.42  & 6.00\% \\  
      &STSGCN          & 1.44  & 3.01  & 3.04\%        & 1.83  & 4.18  & 4.17\%       & 2.26  & 5.21  & 5.40\% \\  
      &MTGNN           & 1.32  & 2.79  & 2.77\%        & 1.65  & 3.74  & 3.69\%       & 1.94  & 4.49  & 4.53\% \\  
      &GMAN            & 1.34  & 2.91  & 2.86\%        & 1.63  & 3.76  & 3.68\%       & 1.86  & 4.32  & \textbf{4.37}\% \\  
      &DGCRN           & 1.28  & 2.69  & 2.66\%        & 1.59  & 3.63  & 3.55\%       & 1.89  & 4.42  & 4.43\% \\  
    \cmidrule(r){2-11}
      &D$^2$STGNN      & \textbf{1.24}$^*$  & \textbf{2.60}$^*$  & \textbf{2.58\%}$^*$        & \textbf{1.55}$^*$  & \textbf{3.52}$^*$  & \textbf{3.49\%}$^*$      & \textbf{1.85}$^*$  & \textbf{4.30}$^*$  & \textbf{4.37\%} \\ 
    \midrule  
    \hline
    \color{black}{\multirow{14}*{\textbf{PEMS04}}}
      &HA              & 28.92  & 42.69  & 20.31\%        & 33.73  & 49.37  & 24.01\%       & 46.97  & 67.43  & 35.11\% \\ 
      &VAR             & 21.94  & 34.30  & 16.42\%        & 23.72  & 36.58  & 18.02\%        & 26.76  & 40.28  & 20.94\% \\ 
      &SVR             & 22.52  & 35.30  & 14.71\%        & 27.63  & 42.23  & 18.29\%       & 37.86  & 56.01  & 26.72\% \\ 
      &FC-LSTM         & 21.42  & 33.37  & 15.32\%        & 25.83  & 39.10  & 20.35\%       & 36.41  & 50.73  & 29.92\% \\ 
      &DCRNN           & 20.34  & 31.94  & 13.65\%        & 23.21  & 36.15  & 15.70\%       & 29.24  & 44.81  & 20.09\% \\ 
      &STGCN           & 19.35  & 30.76  & 12.81\%        & 21.85  & 34.43  & 14.13\%       & 26.97  & 41.11  & 16.84\% \\ 
      &Graph WaveNet   & 18.15  & 29.24  & 12.27\%        & 19.12  & 30.62  & 13.28\%       & 20.69  & 33.02  & 14.11\% \\
      &ASTGCN          & 20.15  & 31.43  & 14.03\%        & 22.09  & 34.34  & 15.47\%       & 26.03  & 40.02  & 19.17\% \\  
      &STSGCN          & 19.41  & 30.69  & 12.82\%        & 21.83  & 34.33  & 14.54\%       & 26.27  & 40.11  & 14.71\% \\  
      &MTGNN           & 18.22  & 30.13  & 12.47\%        & 19.27  & 32.21  & 13.09\%       & 20.93  & 34.49  & 14.02\% \\  
      &GMAN            & 18.28  & 29.32  & 12.35\%        & 18.75  & 30.77  & 12.96\%       & 19.95  & \textbf{30.21}  & 12.97\% \\  
      &DGCRN           & 18.27  & 28.97  & 12.36\%        & 19.39  & 30.86  & 13.42\%       & 21.09  & 33.59  & 14.94\% \\  
    \cmidrule(r){2-11}
      &D$^2$STGNN      & \textbf{17.44}$^*$  & \textbf{28.64}$^*$  & \textbf{11.64\%}$^*$        & \textbf{18.28}$^*$  & \textbf{30.10}$^*$  & \textbf{12.10\%}$^*$      & \textbf{19.55}$^*$  & 31.99  & \textbf{12.82\%}$^*$ \\ 
    
        \midrule
    \hline
    \color{black}{\multirow{14}*{\textbf{PEMS08}}}
      &HA              & 23.52  & 34.96  & 14.72\%        & 27.67  & 40.89  & 17.37\%       & 39.28  & 56.74  & 25.17\% \\ 
      &VAR             & 19.52  & 29.73  & 12.54\%        & 22.25  & 33.30  & 14.23\%        & 26.17  & 38.97  & 17.32\% \\ 
      &SVR             & 17.93  & 27.69  & 10.95\%        & 22.41  & 34.53  & 13.97\%       & 32.11  & 47.03  & 20.99\% \\ 
      &FC-LSTM         & 17.38  & 26.27  & 12.63\%        & 21.22  & 31.97  & 17.32\%       & 30.69  & 43.96  & 25.72\% \\ 
      &DCRNN           & 15.64  & 25.48  & 10.04\%        & 17.88  & 27.63  & 11.38\%       & 22.51  & 34.21  & 14.17\% \\ 
      &STGCN           & 15.30  & 25.03  &  9.88\%        & 17.69  & 27.27  & 11.03\%       & 25.46  & 33.71  & 13.34\% \\ 
      &Graph WaveNet   & 14.02  & 22.76  &  8.95\%        & 15.24  & 24.22  &  9.57\%       & 16.67  & 26.77  & 10.86\% \\
      &ASTGCN          & 16.48  & 25.09  & 11.03\%        & 18.66  & 28.17  & 12.23\%       & 22.83  & 33.68  & 15.24\% \\  
      &STSGCN          & 15.45  & 24.39  & 10.22\%        & 16.93  & 26.53  & 10.84\%       & 19.50  & 30.43  & 12.27\% \\  
      &MTGNN           & 14.24  & 22.43  &  9.02\%        & 15.30  & 24.32  &  9.58\%       & 16.85  & 26.93  & 10.57\% \\  
      &GMAN            & 13.80  & 22.88  &  9.41\%        & 14.62  & 24.02  &  9.57\%       & 15.72  & \textbf{25.96}  & 10.56\% \\  
      &DGCRN           & 13.89  & 22.07  &  9.19\%        & 14.92  & 23.99  & 9.85\%       & 16.73  & 26.88  & 10.84\% \\  
    \cmidrule(r){2-11}
      &D$^2$STGNN      & \textbf{13.14}$^*$  & \textbf{21.42}$^*$  & \textbf{8.55\%}$^*$        & \textbf{14.21}$^*$  & \textbf{23.65}$^*$  & \textbf{9.12\%}$^*$      & \textbf{15.69}$^*$  & 26.41  & \textbf{10.17\%}$^*$ \\ 
        \midrule  
      \bottomrule
    \end{tabular}
\end{table*}
As shown in Table \ref{tab:main}, D$^2$STGNN consistently achieves the best performance in all horizons in all datasets, which indicates the effectiveness of our model.
Traditional methods such as HA, SVR perform worst because of their strong assumption about the data, \eg stationary or linear. 
FC-LSTM, a classic recurrent neural network for sequential data, can not perform well since it only considers temporal features, but ignores the spatial impact in traffic data and, which is crucial in traffic forecasting.
VAR takes both spatial and temporal information into consideration, thus it achieves better performance.
However, VAR cannot capture strong nonlinear and dynamic spatial-temporal correlations. 
{\color{black} Recently proposed spatial-temporal models overcome these shortcomings and make considerable progress. DCRNN and Graph WaveNet are two typical spatial-temporal coupling models among them.}
Graph WaveNet combines GNN and Gated TCN to form a spatial-temporal layer while DCRNN replaces the fully connected layer in GRU by diffusion convolution to get a diffusion convolutional GRU.
Even if compared with many of the latest works, such as ASTGCN and STSGCN, their performance is still very promising. 
This may be due to their refined data assumptions and reasonable model architecture.
MTGNN replaces the GNN and Gated TCN in Graph WaveNet with mix-hop propagation layer~\cite{2019MixHop} and dilated inception layer, and proposes the learning of latent adjacency matrix to seek further improvement.
GMAN performs better in long-term prediction thanks to the attention mechanism's powerful ability to capture long-term dependency.
{\color{black} Based on the DCRNN architecture, DGCRN captures the dynamic characteristics of the spatial topology and achieves better performance than other baselines.
Our model still outperforms DGCRN. We conjecture the key reason lies in the decoupled ST framework.
}
In a nutshell, the results in Table~\ref{tab:main} validate the superiority of D$^2$STGNN.

Note that the final performance is affected by many aspects: the modeling of temporal and spatial dependencies and dynamic spatial topology.
Therefore, although Table \ref{tab:main} has shown the superiority of the D$^2$STGNN model, it is not enough to evaluate the effectiveness of the {\color{black}proposed} decoupled spatial-temporal framework. 

\subsection{Effectiveness of the Decoupled Framework and the Spatial-Temporal Model}
\begin{table}
\renewcommand\arraystretch{0.91}
\caption{Comparison of decoupled and coupled ST Framework. {\color{black}$\mathbf{H}$ denotes horizon}. Numbers marked with $^*$ indicate that the improvement is statistically significant compared with the best baseline~(t-test with p-value$<0.05$).}
\label{tab:dcvsc}
\begin{tabular}{p{0.1cm}<{\centering}p{0.15cm}<{\centering}p{0.8cm}<{\centering}p{0.8cm}<{\centering}<{\centering}p{1cm}<{\centering}p{1.5cm}<{\centering}p{1.5cm}<{\centering}}
\toprule
  \midrule
      & &                               & GWNet     
                                                       & DGCRN$^\dagger$ 
                                                                    &D$^2$STGNN$^\ddagger$ 
                                                                                &D$^2$STGNN$^\dagger$\\
      \midrule
\multirow{9}*{\rotatebox{90}{\textbf{METR-LA}}} 
&\multirow{3}*{\makecell{\textbf{H}\\\textbf{3}}}     
                                & MAE   & 2.69         & 2.71       & 2.66      & \textbf{2.59}$^*$   \\ 
&                               & RMSE  & 5.15         & 5.19       & 5.10      & \textbf{4.99}$^*$    \\ 
&                               & MAPE  & 6.90\%       & 7.04\%     & 6.80\%    & \textbf{6.69\%}$^*$  \\ 
\cmidrule(r){2-7}
&\multirow{3}*{\makecell{\textbf{H}\\\textbf{6}}}     
                                & MAE   & 3.07         & 3.12       & 3.04      & \textbf{2.93}$^*$   \\ 
&                               & RMSE  & 6.22         & 6.31       & 6.13      & \textbf{5.97}$^*$    \\ 
&                               & MAPE  & 8.37\%       & 8.60\%     & 8.24\%    & \textbf{7.99\%}$^*$  \\ 
\cmidrule(r){2-7}
&\multirow{3}*{\makecell{\textbf{H}\\\textbf{12}}}    
                                & MAE   & 3.53         & 3.64       & 3.51      & \textbf{3.38}$^*$   \\ 
&                               & RMSE  & 7.37         & 7.59       & 7.27      & \textbf{7.07}$^*$    \\ 
&                               & MAPE  & 10.01\%      & 10.62\%    & 10.02\%    & \textbf{9.63\%}$^*$  \\    
\midrule
\midrule
\multirow{9}*{\rotatebox{90}{\textbf{PEMS-BAY}}} 
&\multirow{3}*{\makecell{\textbf{H}\\\textbf{3}}}     
                                & MAE   & 1.30         & 1.32        & 1.31      & \textbf{1.25}$^*$   \\ 
&                               & RMSE  & 2.74         & 2.78        & 2.74      & \textbf{2.64}$^*$    \\ 
&                               & MAPE  & 2.73\%       & 2.78\%      & 2.76\%    & \textbf{2.64\%}$^*$  \\ 
\cmidrule(r){2-7}
&\multirow{3}*{\makecell{\textbf{H}\\\textbf{6}}}     
                                & MAE   & 1,63         & 1.66        & 1.63      & \textbf{1.55}$^*$   \\ 
&                               & RMSE  & 3.70         & 3.78        & 3.66      & \textbf{3.56}$^*$    \\ 
&                               & MAPE  & 3.67\%       & 3.76\%      & 3.66\%    & \textbf{3.58\%}$^*$  \\ 
\cmidrule(r){2-7}
&\multirow{3}*{\makecell{\textbf{H}\\\textbf{12}}}    
                                & MAE   & 1.95         & 1.99        & 1.94      & \textbf{1.85}$^*$   \\ 
&                               & RMSE  & 4.52         & 4.60        & 4.50      & \textbf{4.33}$^*$    \\ 
&                               & MAPE  & 4.63\%       & 4.73\%      & 4.59\%    & \textbf{4.43\%}$^*$  \\          

\midrule
\midrule
\multirow{9}*{\rotatebox{90}{\textbf{PEMS04}}}
&\multirow{3}*{\makecell{\textbf{H}\\\textbf{3}}}     
                                & MAE   & 18.15         & 18.97        & 18.94      & \textbf{17.55}$^*$   \\ 
&                               & RMSE  & 29.24         & 30.01        & 29.38      & \textbf{28.70}$^*$    \\ 
&                               & MAPE  & 12.27\%       & 13.38\%      & 13.58\%    & \textbf{11.78\%}$^*$  \\ 
\cmidrule(r){2-7}
&\multirow{3}*{\makecell{\textbf{H}\\\textbf{6}}}     
                                & MAE   & 19.12         & 20.30        & 20.14      & \textbf{18.38}$^*$   \\ 
&                               & RMSE  & 30.62         & 31.78        & 31.54      & \textbf{30.15}$^*$    \\ 
&                               & MAPE  & 13.28\%       & 14.48\%      & 15.11\%    & \textbf{12.26\%}$^*$  \\ 
\cmidrule(r){2-7}
&\multirow{3}*{\makecell{\textbf{H}\\\textbf{12}}} 
                                & MAE   & 20.69        & 22.95         & 22.57      & \textbf{19.59}$^*$   \\ 
&                               & RMSE  & 33.02         & 35.15        & 34.33      & \textbf{32.04}$^*$    \\ 
&                               & MAPE  & 14.11\%       & 16.97\%      & 17.16\%    & \textbf{12.95\%}$^*$  \\      
\midrule
\midrule
\multirow{9}*{\rotatebox{90}{\textbf{PEMS08}}}
&\multirow{3}*{\makecell{\textbf{H}\\\textbf{3}}}     
                                & MAE   & 14.02         & 14.54        & 14.49      & \textbf{13.28}$^*$   \\ 
&                               & RMSE  & 22.76         & 22.62        & 22.35      & \textbf{21.56}$^*$    \\ 
&                               & MAPE  & 8.95\%        & 9.37\%      & 10.14\%    & \textbf{8.53\%}$^*$  \\ 
\cmidrule(r){2-7}
&\multirow{3}*{\makecell{\textbf{H}\\\textbf{6}}}     
                                & MAE   & 15.24         & 15.64        & 15.69      & \textbf{14.26}$^*$   \\ 
&                               & RMSE  & 24.22         & 24.52        & 24.37      & \textbf{23.49}$^*$    \\ 
&                               & MAPE  & 9.57\%        & 10.03\%      & 10.41\%    & \textbf{9.15\%}$^*$  \\ 
\cmidrule(r){2-7}
&\multirow{3}*{\makecell{\textbf{H}\\\textbf{12}}}    
                                & MAE   & 16.67         & 17.80       & 18.01      & \textbf{15.65}$^*$   \\ 
&                               & RMSE  & 26.77         & 27.92        & 27.53      & \textbf{25.78}$^*$    \\ 
&                               & MAPE  & 10.86\%       & 11.71\%      & 11.81\%    & \textbf{10.10\%}$^*$  \\      

\midrule  
\bottomrule
\end{tabular}
\end{table}
In this subsection, we {\color{black}conduct experiments to verify the effectiveness of the decoupled spatial-temporal framework as well as the diffusion and inherent model.}
As mentioned before, we remove the {\color{black}dynamic spatial dependency learning} module in all methods, \eg dynamic graph learner {\color{black}in our model, for a fair comparison}.

On the one hand, we need to compare D$^2$STGNN with its {\color{black}variant without the DSTF, named the coupled version of D$^2$STGNN, where the two hidden time series remain coupled like in other STGNNs.}
On the other hand, we also want to compare the coupled version {\color{black}of D$^2$STGNN} with the other STGNNs to test {\color{black}the effectiveness of the diffusion and inherent model}.
To this end, we first replace the dynamic graph in D$^2$STGNN with the pre-defined static graph to get D$^2$STGNN$^\dagger$.
Based on it, we consider D$^2$STGNN$\ddagger$, which removes the DSTF by removing the {\color{black}estimation} gate and residual decomposition, and connects the {\color{black}diffusion model and inherent model} directly.
We select the two most representative baselines, Graph WaveNet~(GWNet) and DGCRN. 
For a fair comparison, the dynamic adjacency matrix in DGCRN is also removed, named DGCRN$\dagger$.

The result is shown in Table \ref{tab:dcvsc}. 
We have the following findings. 
(i) D$^2$STGNN$^\dagger$ significantly outperforms D$^2$STGNN$^\ddagger$, which shows that the DSTF is crucial in our model.
(ii) The coupled version D$^2$STGNN$^\ddagger$ can also perform better than baselines, which indicates the effectiveness of our {\color{black}diffusion and inherent} model.
However, the D$^2$STGNN$^\ddagger$ has only limited advantages compared with other baselines, {\color{black}which} again shows the importance of {\color{black}decoupling the two hidden time series in the original traffic data.}
\subsection{Efficiency}
\begin{figure}[h]
\setlength{\abovecaptionskip}{0.2cm}
  \centering
  \includegraphics[width=0.95\linewidth]{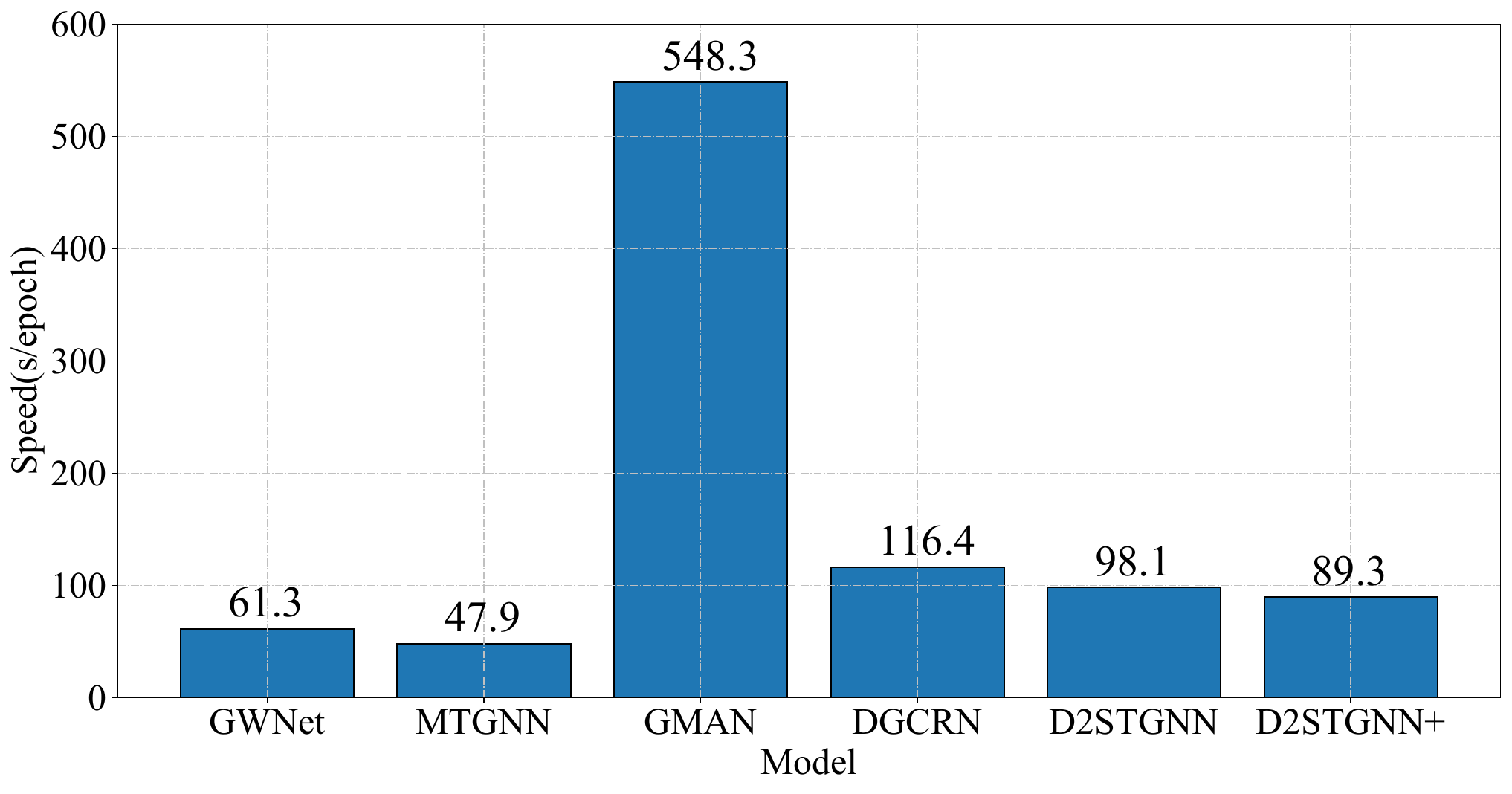}
  \caption{Average training time per epoch.}
  \label{efficiency}
\end{figure}

In this part, we compare the efficiency of D$^2$STGNN with other methods based on the METR-LA dataset.
For a more intuitive and effective comparison, we compare the average training time required for each epoch of these models.
Specifically, we compare the speed of D$^2$STGNN, D$^2$STGNN$^\dagger$~(without dynamic graph learning), DGCRN, GMAN, MTGNN, and Graph WaveNet.
All models are running on Intel(R) Xeon(R) Gold 5217 CPU @ 3.00GHz, 128G RAM computing server, equipped with RTX 3090 graphics card.
The batch size is uniformly set to 32.

As shown in Figure \ref{efficiency}, D$^2$STGNN does not increase the computational burden too much compared to other baselines.
In addition, it achieves both better performance and higher efficiency in the same time than other state-of-the-art baselines like GMAN and DGCRN.
This is mainly due to the fact that the decoupled framework~(\ie  {\color{black}estimation} gate and residual decomposition mechanism) focuses on developing a more reasonable structures connecting {\color{black}diffusion and inherent models} to improve the model's capabilities, rather than increasing the complexity of the {\color{black}diffusion or inherent models}.
Graph WaveNet and MTGNN are the most efficient, thanks to their lightweight and easily parallelized models. But their performance is worse than other models.
\subsection{Ablation Study}
\label{sec_abs}
{
\begin{table*}[ht]
\renewcommand\arraystretch{0.91}
\centering
\caption{Ablation study on METR-LA.}
\label{tab:ablation}
\begin{tabular}{cccr|ccr|ccr}
\toprule
\hline

\multirow{2}*{\textbf{Variants}}    & \multicolumn{3}{c}{\textbf{Horizon 3}}     & \multicolumn{3}{c}{\textbf{Horizon 6}}   & \multicolumn{3}{c}{\textbf{Horizon 12}}\\ 
            \cmidrule(r){2-4} \cmidrule(r){5-7} \cmidrule(r){8-10}
        & MAE & RMSE & MAPE & MAE & RMSE & MAPE & MAE & RMSE & MAPE\\
\midrule
 D$^2$STGNN & \textbf{2.56} & \textbf{4.88} & \textbf{6.48\%}   & \textbf{2.90} & \textbf{5.89} & \textbf{7.78\%}   & \textbf{3.35}  & \textbf{7.03}  & \textbf{9.40\%} \\
 \hline
  \color{black}\textit{switch}     & \color{black}\textbf{2.56}  & \color{black}4.90  & \color{black}6.50\%        & \color{black}2.91  & \color{black}5.92  & \color{black}7.85 \%       & \color{black}\textbf{3.35}  & \color{black}7.05  & \color{black}9.47\%\\
  \color{black}\textit{w/o gate}   & \color{black}2.60  & \color{black}4.98  & \color{black}6.63\%        & \color{black}2.96  & \color{black}6.01  & \color{black}8.07 \%       & \color{black}3.44  & \color{black}7.16  & \color{black}9.78\%\\
  \color{black}\textit{w/o res}    & \color{black}2.60  & \color{black}4.96  & \color{black}6.84\%        & \color{black}2.93  & \color{black}5.95  & \color{black}8.21 \%       & \color{black}3.37  & \color{black}7.10  & \color{black}9.80\%\\
  \color{black}\textit{w/o decouple}   & \color{black}2.66  & \color{black}5.10  & \color{black}6.80\%        & \color{black}3.04  & \color{black}6.13  & \color{black}8.24 \%       & \color{black}3.51  & \color{black}7.27  & \color{black}10.02\%\\
 \hline
  \textit{w/o dg}    & 2.59  & 4.99  & 6.69\%        & 2.93  & 5.97  & 7.99 \%       & 3.38  & 7.07  & 9.63\%\\
  \color{black}\textit{w/o apt}    & \color{black}2.58  & \color{black}4.92  & \color{black}6.51\%        & \color{black}2.93  & \color{black}5.92  & \color{black}7.80 \%       & \color{black}3.40  & \color{black}7.10  & \color{black}9.43\%\\
  \textit{w/o gru}   & 2.59  & 4.93  & 6.66\%        & 2.94  & 6.02  & 7.98 \%       & 3.38  & 7.07  & 9.66\% \\ 
  \textit{w/o msa}   & 2.59  & 4.95  & 6.60\%        & 2.93  & 5.96  & 7.99 \%       & 3.37  & 7.09  & 9.67\% \\ 
 \hline
  \textit{w/o ar}    & 2.59  & 4.98  & 6.61\%        & 2.94  & 5.96  & 7.95 \%       & 3.39  & 7.09  & 9.64\% \\ 
  \textit{w/o cl}    & 2.62  & 5.01  & 6.70\%        & 2.96  & 6.02  & 8.05 \%       & 3.38  & 7.08 & 9.63\% \\ 

\hline
\bottomrule
\end{tabular}
\end{table*}
}
{\color{black}In this part, we will conduct ablation studies from three aspects to verify our work: the architecture of decoupled spatial-temporal framework, the important components, and the training strategy.
Firstly, we design four variants of our decoupled spatial-temporal framework.
\textit{Switch} places inherent block before diffusion block in each layer to verify that whether they are interchangeable. 
\textit{W/o gate} removes the estimation gate in the decouple block, while \textit{w/o res} removes the residual links.
\textit{W/o decouple} removes the estimation gate and residual decomposition simultaneously~(\ie the D$^2$STGNN$\ddagger$ in Table \ref{tab:dcvsc}).
Secondly, we test the effectiveness of four important components.
\textit{W/o dg} replaces the dynamic graph with a pre-defined static graph~(\ie the D$^2$STGNN$\dagger$ in Table \ref{tab:dcvsc}).
\textit{W/o apt} removes the self-adaptive transition matrix in the diffusion model.
\textit{W/o gru} removes the GRU layer in the inherent model, while \textit{w/o msa} removes the multi-head self-attention layer.
Thirdly, we design two variants to test the effectiveness of the training strategy: \textit{w/o ar} removes the auto-regression strategy in the forecast branch and directly applies a regression layer on the hidden state to forecast multi-steps at once, \textit{w/o cl} removes the curriculum learning.

The result is shown in Table \ref{tab:ablation}.
On the architecture aspect, switching the diffusion and inherent model does not make a significant difference.
The results of \textit{w/o gate} and \textit{w/o res} suggest that the estimation gate and residual decomposition mechanism are both important for decoupling.
The results of \textit{w/o decouple} show that decoupling the two hidden signals is crucial for accurate traffic forecasting.
On the important components aspect, the dynamic graph learning model provides consistent performance improvements compared with the pre-defined static graph.
The results of \textit{w/o gru} and \textit{w/o msa} in the inherent model show that both short- and long- term dependencies are crucial for accurate traffic forecasting.
On the training strategy aspect, the result of \textit{w/o ar} indicates that the auto-regressive forecast strategy is more suitable for our model, and the result of \textit{w/o cl} shows that correct training strategy can help the model to converge better.}

\subsection{Parameter Sensitivity}
\begin{figure}[h]
\centering
\setlength{\abovecaptionskip}{0.2cm}
\subfigure[Impact of kernel size $k_t$ and $k_s$]{
\includegraphics[width=0.230\textwidth]{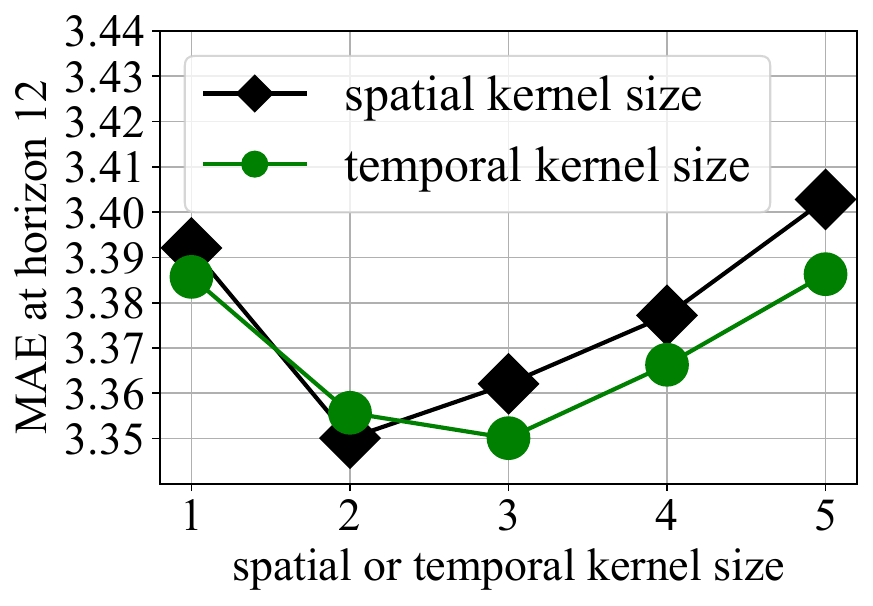}}
\subfigure[Impact of hidden dimension $d$]{
\includegraphics[width=0.230\textwidth]{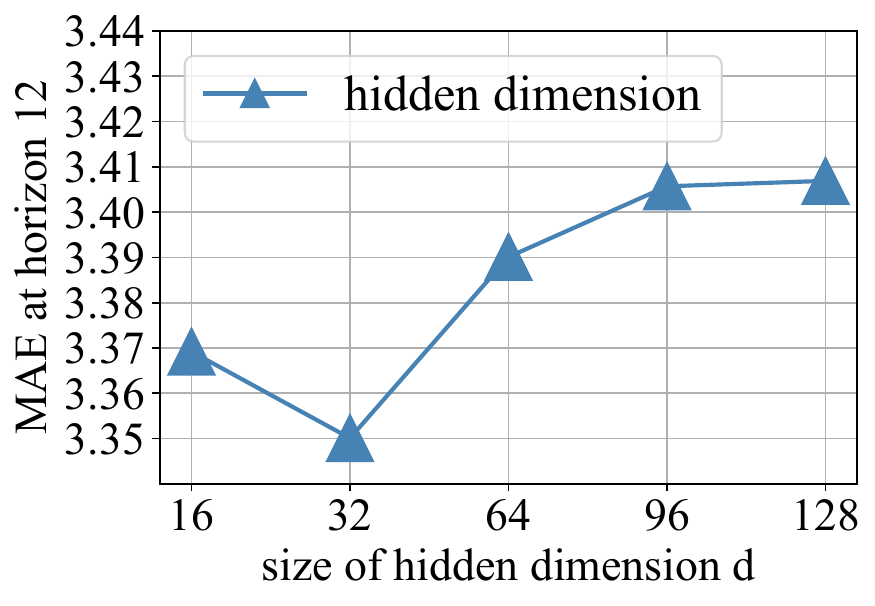}}
\caption{Parameter sensitivity of D$^2$STGNN.}
\label{Parameter}
\end{figure}
In this section, we conduct experiments to analyze the impacts of three critical hyper-parameters: spatial kernel size $k_s$, temporal kernel size $k_t$, and hidden dimension $d$.
In Figure \ref{Parameter}, we present the traffic forecasting result on METR-LA dataset with different parameters.
The effect of spatial kernel size $k_s$ and temporal kernel size $k_t$ is shown in Figure \ref{Parameter}(a).
{\color{black}We set the range of $k_s$ and $k_t$ from 1 to 5 individually.}
The experimental results verify the {\color{black}spatial-temporal} localized characteristics of {\color{black}diffusion process}.
In addition, the effect of hidden dimension $d$ is shown in Figure \ref{Parameter}(b), {\color{black}which shows that} a smaller dimension is insufficient to encode the spatial and temporal information, while the larger dimension may introduce overfitting.

\begin{figure}[h]
\centering 
\setlength{\abovecaptionskip}{0.2cm}
\subfigure[Visualization on node~(sensor) 2]{
\label{Fig.sub.1}
\includegraphics[width=0.48\textwidth]{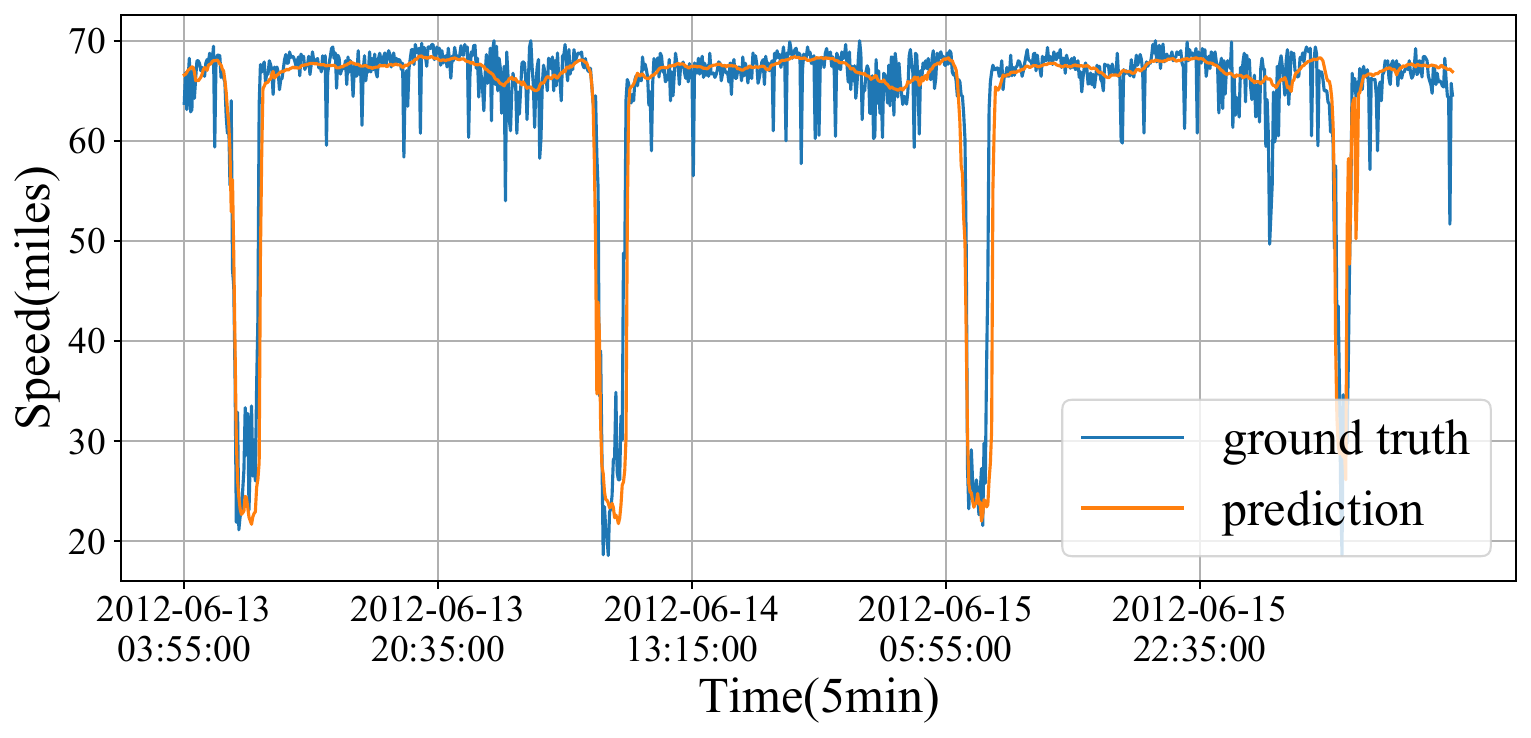}}
\subfigure[Visualization on node~(sensor) 111]{
\label{Fig.sub.2}
\includegraphics[width=0.48\textwidth]{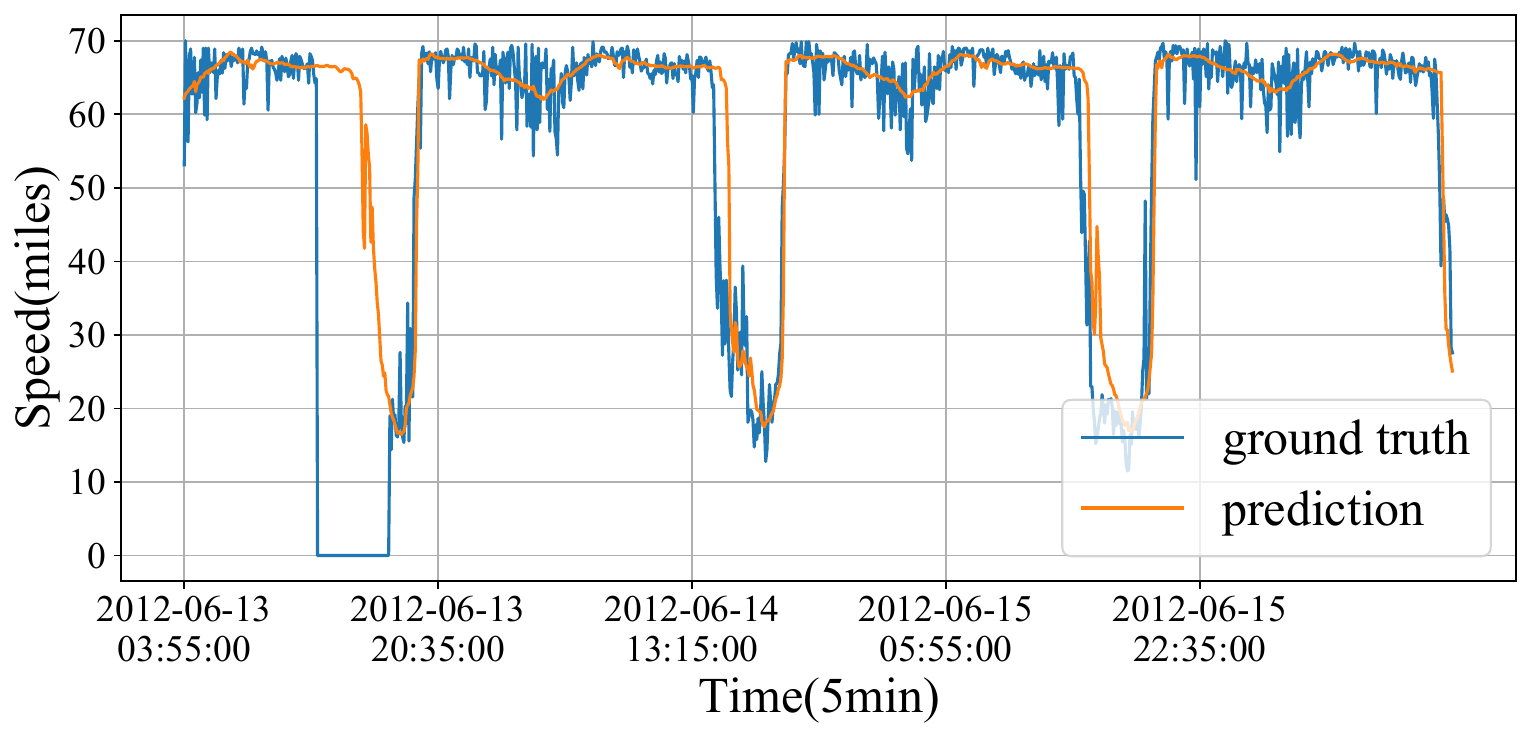}}
\caption{Visualization of prediction results on METR-LA.}
\label{Visualization}
\end{figure}

\subsection{Visualization}

In order to further intuitively understand and evaluate our model, in this section, we visualize the prediction of our model and the {\color{black}ground-truth}.
Due to space limitations, we randomly selected two nodes and displayed their data from June 13th 2012 to June 16th 2012~({\color{black}located} in the test dataset).
The forecasting results on node $2$ and node $111$ are shown in Figure \ref{Visualization}.
It is obvious that the patterns of the two selected nodes are different.
For example, there is often traffic congestion during the morning peak hour at sensor $2$, while sensor $111$ often records traffic congestion during the evening peak hours.
The results indicate that our model can capture these unique patterns for different nodes.
Furthermore, it can be seen that the model is very good at capturing the inherent patterns of time series while avoiding overfitting the noise.
For example, sensor $111$ apparently failed in the afternoon of June 13, 2012, where the records suddenly were zero. 
{\color{black}However,} our model does not forcefully fit these noises and correctly predicted the traffic congestion.
Furthermore, as shown in Figure \ref{Visualization}, the model achieved very impressive prediction accuracy on the whole, while the prediction in some local details may not be accurate due to large random noise.

\section{Conclusion}
\label{Section7}
{\color{black}In this paper, we first propose to decouple the diffusion signal and inherent signal from traffic data by a novel  \textbf{D}ecoupled \textbf{S}patial \textbf{T}emporal \textbf{F}ramework~(DSTF).
This enables more precise modeling of the different parts of traffic data, thus promising to improve prediction accuracy.
Based on the novel DSTF, \textbf{D}ecoupled \textbf{D}ynamic \textbf{S}patial-\textbf{T}emporal \textbf{G}raph \textbf{N}eural \textbf{N}etwork(D$^2$STGNN) is proposed by carefully designing the diffusion and inherent model as well as the dynamic graph learning model according to the characteristics of diffusion signals and inherent signals.
Specifically, a spatial-temporal localized convolution is designed to model \textit{the hidden diffusion time series}. The recurrent neural network and self-attention mechanism are jointly used to model the \textit{hidden inherent time series}.
Furthermore, the dynamic graph learning module comprehensively exploits different information to adjust the road network-based spatial dependency by learning the latent correlations between time series based on the self-attention mechanism.
Extensive experiments on four real-world datasets show that our proposal is able to consistently and significantly outperform all baselines.}

\begin{acks}
This work was supported in part by the National Natural Science Foundation of China under Grant No. 61902376, No. 61902382, and No. 61602197, in part by CCF-AFSG Research Fund under Grant No. RF20210005, and in part by the fund of Joint Laboratory of HUST and Pingan Property \& Casualty Research (HPL).
In addition, Zhao Zhang is supported by the China Postdoctoral Science Foundation under Grant No. 2021M703273.

\end{acks}


\balance 
\bibliographystyle{ACM-Reference-Format}
\normalem
\bibliography{references}

\end{document}